\documentclass[10pt,twocolumn,letterpaper]{article}

\usepackage{cvpr}              
\usepackage{multirow}
\usepackage{multicol}
\usepackage{comment}
\usepackage{sidecap}
\definecolor{cvprblue}{rgb}{0.21,0.49,0.74}
\usepackage[pagebackref,breaklinks,colorlinks,allcolors=cvprblue]{hyperref}
\usepackage{graphicx, booktabs, tabularx}
\usepackage[dvipsnames,svgnames]{xcolor}
\usepackage{comment}
\def\paperID{6643} 
\def\confName{CVPR}
\def\confYear{2026}

\title{\includegraphics[height=1cm]{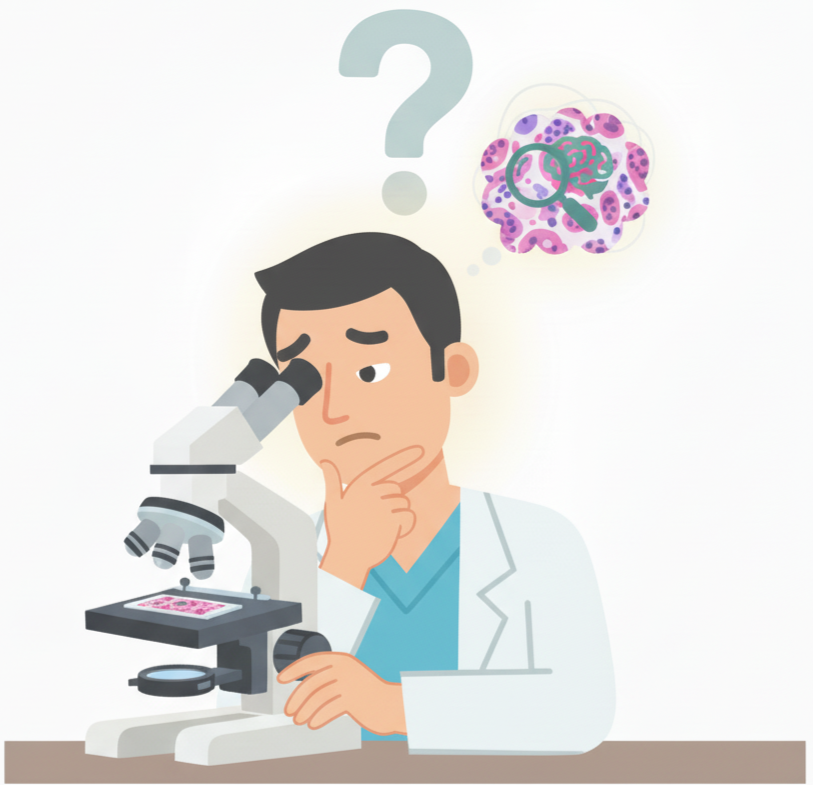}\hspace{0.5em}MLLM-HWSI: A Multimodal Large Language Model for Hierarchical Whole Slide Image Understanding}

\author{Basit Alawode$^{1}$,~Arif Mahmood$^{2}$,~Muaz Khalifa Al-Radi$^{1}$,~Shahad Albastaki$^{1}$,\\~Asim Khan$^{1}$,~Muhammad Bilal$^{3}$,~Moshira Ali Abdalla$^{1}$,~Mohammed Bennamoun$^{4}$,~Sajid Javed$^{1}$\\$^{1}$Department of Computer Science, Khalifa University of Science and Technology, UAE.\\$^{2}$Information Technology University, Pakistan.$^{3}$KAU, KSA. ~$^{4}$University of the Western Australia.\\
}

\begin{document}
\maketitle

\begin{abstract}
Whole Slide Images (WSIs) exhibit hierarchical structure, where diagnostic information emerges from cellular morphology, regional tissue organization, and global context.
Existing Computational Pathology (CPath) Multimodal Large Language Models (MLLMs) typically compress an entire WSI into a single embedding, which hinders fine-grained grounding and ignores how pathologists synthesize evidence across different scales. 
We introduce \textbf{MLLM-HWSI}, a Hierarchical WSI-level MLLM that aligns visual features with pathology language at four distinct scales, cell as word, patch as phrase, region as sentence, and WSI as paragraph to support interpretable evidence-grounded reasoning. 
MLLM-HWSI decomposes each WSI into multi-scale embeddings with scale-specific projectors and jointly enforces (i) a hierarchical contrastive objective and (ii) a cross-scale consistency loss, preserving semantic coherence from cells to the WSI.
We compute diagnostically relevant patches and aggregate segmented cell embeddings into a compact cellular token per-patch using a lightweight \textit{Cell–Cell Attention Fusion (CCAF)} transformer.
The projected multi-scale tokens are fused with text tokens and fed to an instruction-tuned LLM for open-ended reasoning, VQA, report, and caption generation tasks. 
Trained in three stages, MLLM-HWSI achieves new SOTA results on 13 WSI-level benchmarks across six CPath tasks. 
By aligning language with multi-scale visual evidence, MLLM-HWSI provides accurate, interpretable outputs that mirror diagnostic workflows and advance holistic WSI understanding. 
Code is available at: \href{https://github.com/BasitAlawode/HWSI-MLLM}{GitHub}.
\end{abstract}

\vspace{-.7cm}
\begin{figure}
\centering
        \includegraphics[width=3in]{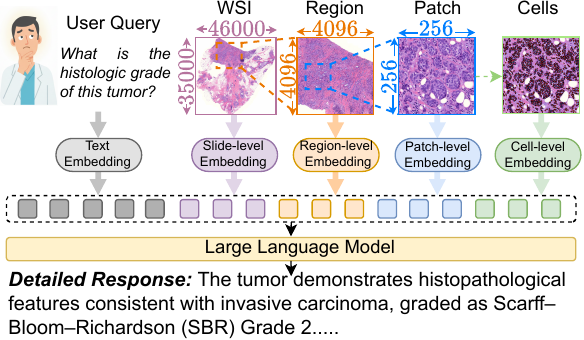} 
\vspace{-2mm}
\caption{Our proposed \textbf{MLLM-HWSI} model aligns WSIs across multiple scales e.g., cells, patches, regions, and WSI enabling fine-grained, context-aware, and interpretable pathology reasoning.}
\label{fig1}
\vspace{-2em}
\end{figure}

\section{Introduction}
\label{sec:intro}
Cancer diagnosis and prognosis using gigapixel Whole Slide Images (WSIs) remain the clinical gold standard for histopathological assessment \cite{wang2020computational, chan2014wonderful, pantanowitz2011review, pantanowitz2013validating, zarella2019practical}. 
The rise of Computational Pathology (CPath) has opened new possibilities to accelerate diagnostic workflows, improve reproducibility, and enable earlier cancer detection through quantitative analysis of the histology landscape \cite{cui2021artificial, song2023artificial, fuchs2011computational}. 
\textit{WSIs are inherently hierarchical, both biologically and structurally, capturing the full spatial organization of tissue across multiple magnifications and scales (Fig. \ref{fig1})\cite{cornish2012whole, chen2022scaling, zarella2019practical, hanna2020whole}. }
This hierarchical organization reflects the architecture of tissue itself, where diagnostic cues emerge across nested levels, from cellular morphology to regional, and global structural patterns \cite{boyce2015whole, chen2022scaling, zarella2019practical, hanna2020whole}.
At the \textbf{cellular level}, WSIs capture diverse morphological attributes including variations in nuclear size, cytoplasmic texture, and mitotic activity that collectively define the vocabulary of pathology \cite{parwani2022whole, mcclintock2021whole, ardon2025digital}. 
At the \textbf{regional level}, these cells form micro-architectural structures such as glands, ducts, or solid nests, which define the syntax of tissue organization and carry diagnostic meaning \cite{brixtel2022whole, nan2025deep}. 
At the \textbf{global WSI level}, multiple regions integrate into a coherent tissue architecture, illustrating spatial relationships between tumor and normal areas, invasion of adjacent structures, and necrosis \cite{tran2025navigating, brixtel2022whole, chen2022scaling, baidoshvili2023whole}. 
\textit{This multiscale organization forms the biological foundation of histopathologic interpretation, underpinning how both human experts and computational models reason about cancer \cite{baidoshvili2023whole,jahn2020digital}.}
\textbf{Expert pathologists} perceive a WSI not as a static but as a multiscale landscape \cite{fraggetta2017routine, brixtel2022whole, baidoshvili2023whole, hijazi2024digital}. 
Diagnostic reasoning typically begins at low magnification, progresses to the examination of regional tissue morphology, and concludes in the inspection of cellular features \cite{plass2023explainability, ramamurthy2015perspective}.
Pathologists interpret WSIs as structured narratives in which tissue architecture provides context, regions define syntax, and cells define vocabulary \cite{kiran2023digital, betmouni2021diagnostic, dimitriou2024magnifying}.
This process is bidirectional: global context informs local inspection, while local findings refine global understanding until a coherent finding is reached \cite{fraggetta2017routine, brixtel2022whole}. 

In CPath, Multimodal Large Language Models (MLLMs) including Quilt-LLaVA \cite{seyfioglu2024quilt}, SlideChat \cite{chen2025slidechat}, WSI-LLaVA \cite{liang2025wsillavamultimodallargelanguage}, TITAN \cite{ding2024titan}, PRISM \cite{shaikovski2024prism}, and HistGen \cite{guo2024histgen} have been proposed for a wide range of tasks, such as Visual Question Answering (VQA), morphological reasoning, and report generation \cite{chen2025slidechat, seyfioglu2024quilt}.
SOTA MLLMs such as SlideChat \cite{chen2025slidechat} and WSI-LLaVA \cite{liang2025wsillavamultimodallargelanguage}, \textit{aggregate patch-level embeddings into a single WSI-level representation aligned with corresponding reports \cite{guo2024histgen, ding2024titan}. }
\textit{Although this aggregation captures a higher-level context, it neglects the hierarchical composition of WSIs, leading to the loss of fine-grained spatial semantics \cite{chen2025slidechat, chen2025wsi, Liang2024_WSI-LLaVA}. }
Also, existing models overlook the clinical workflow of expert pathologists, who integrate multi-scale visual cues obtained from progressive zooming and contextual reasoning \cite{ardon2025digital, ghaznavi2013digital}.

\begin{figure}[t!]
   \centering 
    \includegraphics[width=3in]{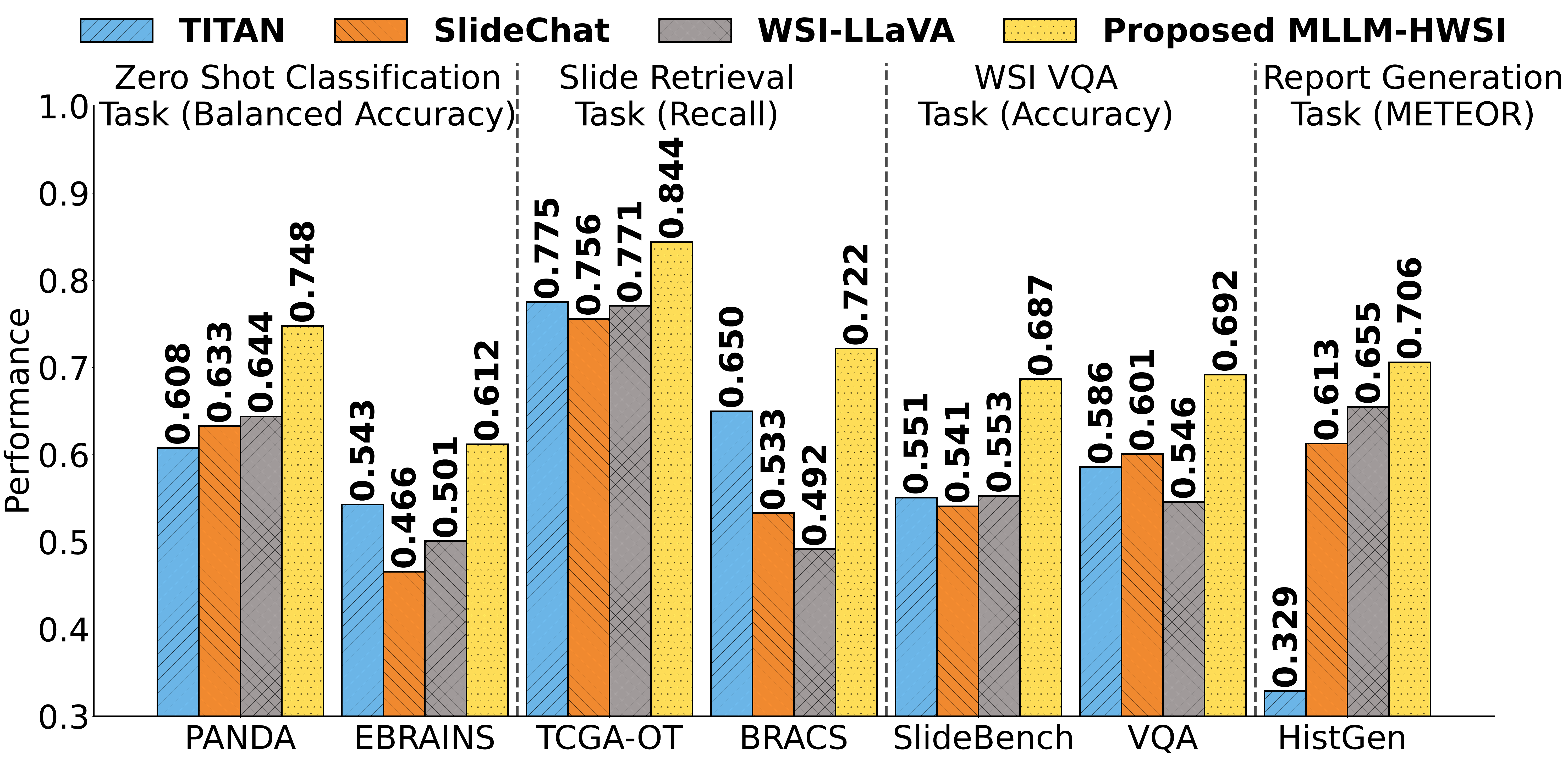} 
     \caption{Comparison of MLLM-HWSI with SOTA methods.}
  \label{fig2}    
\vspace{-2em}
\end{figure}

\textit{In this work, we address these limitations by introducing a Hierarchical WSI-level MLLM (MLLM-HWSI) for comprehensive WSI understanding, including analysis, retrieval, pathological inference, and report generation (Figs. \ref{fig1}-\ref{fig2}).} 
Our approach decodes the inherent \textit{pathology language} by interpreting individual \textit{cells as words}, \textit{small patches as phrases} that describe cellular neighborhoods, \textit{larger regions as sentences} that depict tissue architecture, and the \textit{entire WSI as a paragraph} that forms a coherent visual narrative of the disease \cite{steele2020vocabulary, cheville1983cell, culling2014cellular}. 
We align the hierarchical structure of WSIs with pathology reports across multiple scales, ensuring that MLLM-HWSI mimics the standard diagnostic workflow of pathologists. 
By grounding textual description (e.g., pleomorphic nuclei, stromal invasion) in their corresponding visual counterparts, the model captures compositional reasoning underlying expert diagnosis. 
This multi-scale alignment enhances interpretability, enabling biologically grounded and explainable predictions (Fig. \ref{fig2}).
MLLM-HWSI bridges the gap between tissue-level interpretation by pathologists and computational model reasoning. 
Unlike SlideChat \cite{chen2025slidechat}, TITAN \cite{ding2024titan}, and WSI-LLaVA \cite{liang2025wsillavamultimodallargelanguage}, which rely solely on global embeddings, our model decomposes each WSI into multiple semantic scales-cells, patches, regions, and global WSI—and learns distinct representations for each (Fig. \ref{fig1}).
At the cellular scale, segmented cells are embedded to represent morphological and cytoplasmic features, and a lightweight Vision Transformer (ViT) with a Cell–Cell cross-Embedding Fusion (CCEF) module aggregates cellular information efficiently. 
At higher scales, a hierarchical encoder extracts patch, region, and WSI-level embeddings representing local tissue structure and global architecture. 
A Semantic Patch Filtering module further refines patch-level tokens. 
These embeddings are projected into a shared multimodal space through scale-specific Vision–Language (VL) projectors and aligned with corresponding textual descriptions. 
\textit{By jointly enforcing hierarchical alignment and cross-scale consistency, MLLM-HWSI preserves diagnostic relationships between local cellular features and global structural patterns. }
Aligned visual tokens are then fused with textual tokens during LLM pretraining, enabling multi-scale, evidence-based reasoning.

\noindent MLLM-HWSI is optimized via a hierarchical contrastive alignment loss and a cross-scale consistency loss to maintain semantic coherence across spatial hierarchies. 
Finally, the fused multi-scale visual and textual tokens pre-train an LLM capable of multi-scale interpretative reasoning, mirroring how pathologists integrate detail and context into coherent diagnoses.
We evaluate our proposed MLLM-HWSI model on six different WSI-level CPath tasks including zero-shot classification, retrieval, VQA, report generation, captioning, and cross-modal retrieval using 13 publicly available datasets. 
\textit{Compared to 24 SOTA CPath models, MLLM-HWSI achieves substantial performance improvements as shown in Fig. \ref{fig2}.}
\noindent \textbf{Our main contributions are:}

\begin{enumerate}
    \item We introduce a multi-scale hierarchical MLLM that performs cell-, patch-, region-, and WSI-level alignment with pathology reports, enabling unified multi-scale understanding and reasoning over WSIs.
    \item We jointly optimize hierarchical contrastive alignment and cross-scale consistency losses to preserve semantic coherence across scales, enabling multi-scale and evidence-based reasoning.
    \item By unifying visual hierarchies with pathology reports, our model enhances diagnostic accuracy and generalization compared to global-only MLLMs.
\end{enumerate}
\vspace{-1em}

\section{Literature Review}
\label{sec:review}
\textbf{1. MLLMs in CPath:} MLLMs integrate LLMs with visual encoders to perform instruction-following, reasoning, and report-generation tasks in CPath \cite{chen2025wsi, seyfioglu2024quilt}. 
By coupling visual representations with powerful LLMs (e.g., GPT or LLAMA), these models generate pathology reports, answer clinical queries, and explain diagnostic findings in natural language. 
Patch-level MLLMs such as Quilt-LLaVA \cite{seyfioglu2024quilt} extend VLM pretraining to interactive dialogue and captioning. 
Similarly, WSI-level MLLMs such as PathChat \cite{lu2024multimodal}, TITAN \cite{ding2024titan}, SlideChat \cite{chen2025slidechat}, and WSI-LLaVA \cite{liang2025wsillavamultimodallargelanguage} enable open-ended reasoning across WSIs \cite{seyfioglu2024quilt}. 
\textit{However, most existing CPath MLLMs rely on global WSI-level embeddings that compress the entire WSI into a single vector aligned with a full pathology report.} 
While effective for coarse-level reasoning, this approach neglects the multiscale, hierarchical nature of pathology, limiting the model’s ability to associate textual descriptions with localized visual evidence (Fig. \ref{fig2}). 
\textit{Our Hierarchical WSI-level MLLM (MLLM-HWSI) addresses this gap by aligning features across multiple scales—cell, patch, region, and WSI—with corresponding pathology vocabulary in diagnostic reports, enabling interpretable and biologically grounded reasoning}. \noindent \textbf{2. VLMs in CPath:}
CPath VLMs align histology patches with pathology-specific descriptions, producing semantically meaningful visual representations \cite{lu2023visual, huang2023visual}.
Several prominent VLMs including CONCH \cite{lu2024visual}, PLIP \cite{huang2023visual}, QuiltNet \cite{ikezogwo2024quilt}, CPLIP \cite{javed2024cplip}, MR-PLIP \cite{albastaki2025multi}, and OmniPath \cite{sun2025cpath} have demonstrated improved performance across diverse pathology-related tasks.
The patch-level embeddings from these VLMs are typically aggregated into global representations for WSI-level tasks.
\textit{However, SOTA VLMs primarily operate at the patch-level and fail to explicitly capture the hierarchical organization of WSIs, where diagnostic insights arise from cellular, regional, and global structures.}
\noindent \textbf{3. Visual Foundation Models in CPath:} These models are pretrained on large-scale pathology datasets using a self-supervised learning paradigm \cite{chen2024towards, kang2023benchmarking, wang2022transformer}. 
These models learn transferable, general-purpose visual representations applicable to diverse downstream tasks, including classification and survival prediction \cite{chen2024towards}. 
Prominent patch-level models are CTransPath \cite{wang2022transformer}, UNI \cite{chen2024towards}, DINOSSLPath \cite{kang2023benchmarking}, Virchow \cite{vorontsov2024foundation}, Phikon \cite{filiot2024phikon}, CHIEF \cite{wang2024pathology}, GigaPath \cite{xu2024whole}, and REMEDIS \cite{azizi2022robust}. 
These models act as powerful visual feature extractors capable of encoding cellular and subcellular morphology with strong generalization across tissue types and cancer cohorts \cite{mahmood2024tcgaot}.
At the WSI-level, these models aggregate local patch-level representation popular examples are GigaPath \cite{xu2024whole} and Virchow2 \cite{zimmermann2024virchow2}. 
Such models serve as the visual backbone of modern CPath, offering scalable and generalizable representations for both discriminative and generative pathology tasks. 
\textit{In our work, we adopt these backbones as hierarchical encoders to extract multi-scale WSI features.}


\vspace{-.3cm}
\section{Proposed Hierarchical WSI MLLM}
\label{sec:method}
\noindent \textbf{Overview:} In this work, we propose Hierarchical WSI-level Multimodal Large Language Model (MLLM-HWSI), a unified framework for multi-scale visual understanding and language alignment of WSIs in CPath. 
MLLM-HWSI aims to align the textual content of a pathology report with specific spatial and morphological features within a WSI, ranging from fine-grained cellular morphology to global tissue organization.  
By aligning hierarchical visual-textual representation, MLLM-HWSI enables interpretable, coherent diagnostic reasoning that parallels how pathologists integrate observations across hierarchical scales.

\begin{figure*}[t]
\centering
\includegraphics[width=7.0in,height=4.4in]{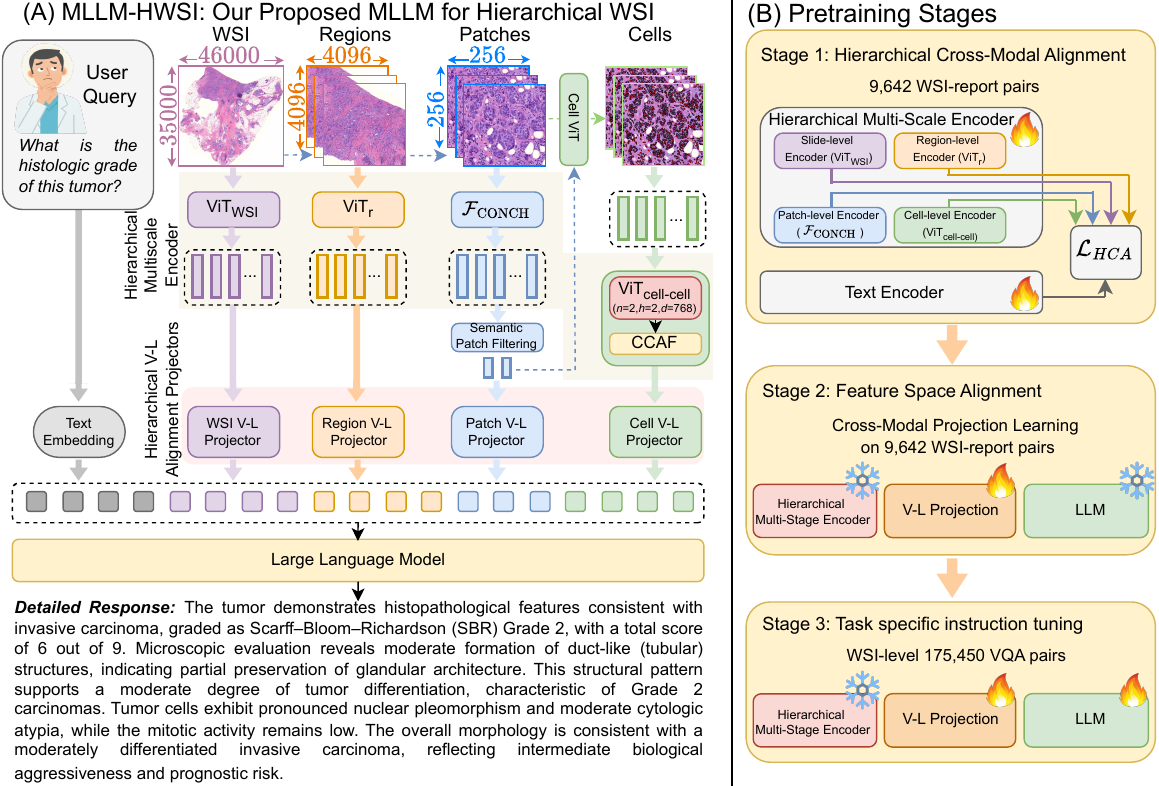}
\vspace{-6mm}
\caption{\textbf{Overview of the proposed MLLM-HWSI.} 
(A) Hierarchical decomposition of WSI into cell, patch, and region-level embeddings aligned with MLLM.
(B) MLLM-HWSI three stage pre-training paradigm for multimodal reasoning.}
\label{fig3}
\end{figure*}

An overview of MLLM-HWSI architecture is illustrated in Fig. \ref{fig3} (A). 
It employs a hierarchical multi-encoder design to capture semantic information at four hierarchical levels. 
At the cellular scale, a CellViT encoder \cite{horst2024cellvit} performs cell segmentation and extracts cell-level embeddings that describe nuclear morphology. 
Three additional encoders process patch, region, and WSI-level representations to capture progressively broader structural and contextual information. 
To efficiently process WSIs, we introduce two key modules: \textbf{Semantic Patch Filtering (SPF)} and \textbf{Cell–Cell Attention Fusion (CCAF)}. 
SPF removes homogeneous patches and selects diagnostically meaningful heterogeneous ones based on cosine similarity with textual embeddings, for multimodal pretraining. 
CCAF employs a lightweight ViT that performs cross-attention among cellular embeddings within each patch, producing a single aggregated cellular token that captures cell-level morphology.

At each hierarchical level, the resulting embeddings are projected into a shared multimodal space using scale-specific VL projectors that align visual features with corresponding textual semantics from pathology reports. 
MLLM-HWSI jointly optimizes two complementary objectives: 
(1) a \textbf{hierarchical contrastive alignment loss}, which strengthens cross-modal correspondence between textual and visual features at each scale, and 
(2) a \textbf{cross-scale consistency loss}, which enforces semantic coherence and hierarchical alignment across different spatial levels.
For multimodal reasoning, the aligned multi-scale embeddings are fused with textual tokens and integrated into an LLM, enabling hierarchical instruction tuning. 
During pretraining, both VL projectors and multi-scale encoder are optimized jointly, achieving end-to-end VL alignment across scales. 

\subsection{Hierarchical Decomposition of Gigapixel WSIs}
WSIs often exceed $100,{000} \times 100,{000}$ pixels, thus direct end-to-end processing is computationally infeasible.
We perform hierarchical decomposition of WSIs to efficiently capture both fine-grained cellular morphology and global tissue context \cite{chen2022scaling}. 
This not only mitigates the processing challenge but also reflects the pathologists' workflow.

In our model, WSI $I$ at 20$\times$ is divided into non-overlapping regions, $I = \{ R_i \}_{i=1}^{n_r},~R_i \in \mathbb{R}^{4096 \times 4096 \times 3}$, where each region $R_i$ preserves sufficient mesoscopic context to capture tissue organization patterns.  
Each region is further subdivided into smaller patches, $ R_i = \{ P_{ij} \}_{j=1}^{n_p},~P_{ij} \in \mathbb{R}^{256 \times 256 \times 3}$. 
In total, we extracted \textit{0.356M regions} and \textit{91.33M patches} from \textit{9,642 WSIs.}
\noindent Hierarchical decomposition allows efficient multi-scale feature extraction while maintaining spatial correspondence across levels. 
It also enables MLLM-HWSI to integrate information from $\{ P_{ij} \}_{j=1}^{n_p} \rightarrow \{ R_i \}_{i=1}^{n_r} \rightarrow I$, facilitating hierarchical VL alignment.
\subsection{Architecture}
The overall architecture of the proposed MLLM-HWSI comprises five key components (Fig. \ref{fig3} (A)): 
(i) a \textbf{Hierarchical Multi-Scale Encoder}, 
(ii) a \textbf{Cell–Cell Attention Fusion (CCAF)} module, 
(iii) a \textbf{Semantic Patch Filtering (SPF)} mechanism, 
(iv) \textbf{Hierarchical V$\rightarrow$L Alignment Projectors}, and 
(v) a \textbf{LLM}. 
Together, these components enable MLLM-HWSI for robust multimodal reasoning. 

\subsection{Hierarchical Multi-Scale Encoder}
\label{sec:encoder}
The hierarchical encoder captures WSI semantics across four spatial levels—cell, patch, region, and WSI, reflecting the diagnostic reasoning process of expert pathologists.

\noindent \textbf{Patch-Level Encoder:} 
At the patch level, visual embeddings are extracted using the CONCH encoder \cite{lu2024visual}, which captures fine-grained texture and mesoscopic structural cues such as glandular formation and stromal organization:
$f_{ij} = \mathcal{F}_{\textrm{CONCH}}(P_{ij})$, where $f_{ij}\in \mathbb{R}^{d_p}$ denotes the representation of patch $P_{ij}$.


\noindent \textbf{Semantic Patch Filtering (SPF):}  
Given the large number of patches $\{ P_{ij} \}_{j=1}^{n_p}$ in a WSI, SPF is introduced to remove redundant and homogeneous patches while retaining diagnostically diverse and report-relevant ones.  
For each region $R_i$, the corresponding patch embeddings $\{ f_{ij} \}_{j=1}^{n_p}$ are normalized, and pairwise cosine similarity is computed as:
\vspace{-3mm}
\begin{equation}
\hat{f}_{ij}= \frac{f_{ij}}{\|f_{ij}\|_2},~s^{j,k}_{i} = \hat{f}_{ij} \cdot \hat{f}_{ik},~\tau_i = \mu_i + \sigma_i,
\vspace{-3mm}
\end{equation}
where $\mu_i = \frac{1}{n_{p}^{2}} \sum_j \sum_k s^{j,k}_i$ is the mean similarity, and $\sigma_i^2 = \frac{1}{n_p^2} \sum_j \sum_k (s^{j,k}_i - \mu_i)^2$ denotes the variance of similarity scores within $R_i$.  
$P_{ij}$ is considered redundant if its mean similarity $\mu^j_i = \frac{1}{n_p} \sum_{k=1}^{n_p} s^{j,k}_i > \tau_i$; otherwise, it is retained in the subset $R_{i}^{'} = \{ P_{ij} \}_{j=1}^{h_{i}}$, where $h_i < n_p$.

\noindent Next, to identify diagnostically relevant patches, the pathology report ($D$) is tokenized into $M$ semantic entities: $D = \{ w_1, w_2, \ldots, w_M \}$ \cite{alsentzer2019publicly}.  
Each token $w_m$ is encoded via the CONCH text encoder $\mathcal{T}_{\textrm{CONCH}}$:
\vspace{-2mm}
\begin{equation}
\mathbf{t}_m = \mathcal{T}_{\textrm{CONCH}}(w_m), \quad
\hat{\mathbf{t}}_m = \frac{\mathbf{t}_m}{\|\mathbf{t}_m\|_2}, 
\quad m \in \{1, \ldots, M\}.
\vspace{-2mm}
\end{equation}

\noindent Cosine similarity between each patch embedding and keyword embedding is then computed as: $s_{ij,m} = \hat{f}_{ij}^{\top}\hat{\mathbf{t}}_m$.
The overall relevance of each patch is quantified by: $r_{ij}= \frac{1}{M}\sum_{m=1}^{M}s_{ij,m}$.
Finally, the top-$k$ patches with the highest relevance scores are selected: $P_{ij} \in R_{i}^{'} ~|~\text{rank}(r_{ij}) \le k$.
The resulting subset $\hat{R}_{i}$ forms a compact, semantically aligned representation with pathology keywords.


\noindent \textbf{Cell-Level Encoder:}  
At cellular scale, each patch $P_{ij} \in \hat{R}_{i}$ is processed by the CellViT encoder \cite{horst2024cellvit}, which performs cell segmentation and encodes nuclear morphology:
\vspace{-2mm}
\begin{equation}
\{c_{ijk}\}_{k=1}^{n_{ij}}= \texttt{CellViT}(P_{ij}),~~\forall P_{ij} \in \hat{R}_{i},
\vspace{-2mm}
\end{equation}
where $c_{ijk} \in \mathbb{R}^{d_{c}}$ represents the embedding of cell $k$ within patch $P_{ij}$, and $n_{ij}$ is the number of segmented cells.  
Given the large number of cells (often exceeding 100K per WSI), we introduce a \textbf{Cell–Cell Attention Fusion (CCAF)} module to aggregate cell-level embeddings efficiently.  
CCAF employs a lightweight ViT that performs cross-attention among $\{c_{ijk}\}$ within each patch, producing a compact token $c_{ij}$ summarizing cell–cell interactions:
\vspace{-2mm}
\begin{equation}
    c_{ij}= \textrm{ViT}_{\textrm{cell-cell}} ([\textrm{CLS}]_{ij},\{c_{ijk}\}_{k=1}^{n_{ij}}), \quad c_{ij} \in \mathbb{R}^{784},
\vspace{-2mm}
\end{equation}
\noindent where $[\textrm{CLS}]_{ij}$ is the token appended next to the sequence $\{c_{ijk} \}_{k=1}^{n_{ij}}$ in $\textrm{ViT}_{\textrm{cell-cell}}$.
This operation yields a single cellular descriptor per patch, encapsulating nuclear diversity and intra-patch morphological context.

\noindent \textbf{Region-Level Encoder:}  
At region-level, we adopt the HIPT hierarchical encoder \cite{chen2022scaling}, denoted as $\textrm{ViT}_{r}$, which aggregates patch-level representations ($p_{ij}$) using $\textrm{ViT}_{p}$ into region-level embeddings that encode micro-architectural dependencies such as tissue polarity, glandular organization, and stromal invasion: $p_{ij} = \textrm{ViT}_{p} (\{ P_{ij} \}_{j=1}^{256}), r_i= \textrm{ViT}_{r}(\{ p_{ij} \}_{j=1}^{256})$.
The resulting $r_i$ provides mesoscopic abstraction bridging cellular features and global context.

\noindent \textbf{WSI-Level Encoder:}  The WSI-level encoder integrates region embeddings $\{ r_i \}_{i=1}^{n_r}$ into a global representation that captures WSI-level wide histological patterns such as tumor distribution: $f_{\textrm{WSI}}= \textrm{ViT}_{WSI} (\{ r_{i}\}_{i=1}^{n_{r}})$.
The $\textrm{ViT}_{WSI}$ architecture follows HIPT \cite{chen2022scaling} but is pre-trained to enhance global tissue-level representation learning.

\noindent \textbf{Final Hierarchical Representation:}  
The resulting multi-scale representation of a WSI is expressed as:
\vspace{-3mm}
\begin{equation}
\mathbf{F}_{\textrm{WSI}} =  \{ \{ c_{ij}, f_{ij}\}_{j=1}^{h_{i}}, r_{i}\}_{i=1}^{n_{r}}, f_{\textrm{WSI}} \}.
\vspace{-2mm}
\end{equation}
This hierarchical structure enables MLLM-HWSI to jointly model cellular morphology, regional organization, and global tissue architecture—providing a biologically VL alignment and diagnostic reasoning.
\subsection{Hierarchical Alignment (V $\rightarrow$ L) Projectors}
To align hierarchical visual features with the language model’s latent space, we employ four distinct V$\rightarrow$L projectors corresponding to each scale: 
cell-level ($A_{c}$), patch-level ($A_{p}$), region-level ($A_{r}$), and WSI-level ($A_{\textrm{WSI}}$). 
The projected features at each level are expressed as:  $z_{c}=A_{c}(c_{ij}), z_{p}=A_{p}(f_{ij}), z_{r}=A_{r}(r_{i}), z_{\textrm{WSI}}=A_{\textrm{WSI}}(f_{\textrm{WSI}})$.

\subsection{Multimodal Large Language Model (LLM)}
The projected embeddings are concatenated with tokenized textual instruction embeddings $z_{\textrm{text}} \in \mathbb{R}^{l \times d_{t}}$ to form the final multimodal input sequence:
$Z = [z_{c}, z_{p}, z_{r}, z_{\textrm{WSI}}, z_{\textrm{text}}]$,  which is then fed into the LLM.
This fusion enables MLLM-HWSI to reason jointly over cell $\rightarrow$ patch $\rightarrow$ region $\rightarrow$ WSI, and the textual context, allowing comprehensive diagnostic interpretation. 
We adopt Qwen2.5-7B-Instruct \cite{yang2024qwen2} as a backbone LLM due to its strong reasoning and instruction-following capabilities.

\begin{table*}[t!]
\begin{minipage}{0.70\textwidth}
   \centering \small
  \setlength{\tabcolsep}{3.0pt}
\centering
\scalebox{0.92}{
\begin{tabular}{l|llcc|cccc}
\hline
\multirow{2}{*}{Models}&Cell&Patch&Region&WSI&PANDA& EBRAINS &WSI-VQA& SlideBench-VQA\\
&Feat.&Feat.&Feat.&Feat.&\cite{Bulten2022_PANDA} (BA)&\cite{roetzer2022digital} (BA)&\cite{chen2025wsi} (A)&(BCNB) \cite{chen2025slidechat} (A)\\
\hline
WSI-LLaVA \cite{liang2025wsillavamultimodallargelanguage}&$\times$&$\times$&$\times$&$\checkmark$&0.644&0.501&0.546&0.553\\
SlideChat \cite{chen2025slidechat}&$\times$&$\times$&$\times$&$\checkmark$&0.633&0.466&0.601&0.541\\
\hline
MLLM-HWSI$_{1}$&$\times$&$\times$&$\times$&$\checkmark$&0.661&0.519&0.616&0.576\\
MLLM-HWSI$_{2}$&$\times$&$\times$&$\checkmark$&$\checkmark$&0.686&0.534&0.611&0.592\\
MLLM-HWSI$_{3}$&$\times$&$\checkmark$&$\checkmark$&$\checkmark$&0.711&0.566&0.661&0.621\\
MLLM-HWSI$_{4}$&$\checkmark$&$\times$&$\times$&$\checkmark$&0.674&0.531&0.613&0.588\\
MLLM-HWSI$_{5}$&$\times$&$\checkmark$&$\times$&$\checkmark$&0.698&0.548&0.623&0.606\\
MLLM-HWSI$_{6}$&$\checkmark$&$\checkmark$&$\times$&$\checkmark$&\underline{0.715}&0.575&\underline{0.669}&0.640\\
MLLM-HWSI$_{7}$&$\checkmark$&$\times$&$\checkmark$&$\checkmark$&0.714&\underline{0.587}&0.668&\underline{0.653}\\
MLLM-HWSI&$\checkmark$&$\checkmark$&$\checkmark$&$\checkmark$&\textbf{0.748}&\textbf{0.612}&\textbf{0.692}&\textbf{0.687}\\
\hline
\end{tabular}
}
\end{minipage}
  \hfill
  \begin{minipage}{0.28\textwidth}
  \small \vspace{1em}
\caption{
\textbf{Ablation 1: Effect of hierarchical representations in MLLM-HWSI.}
Progressive inclusion of cell-, patch-, region-, and WSI-level features improves performance across all benchmarks.
The full MLLM-HWSI achieves the highest scores, confirming the importance of hierarchical multi-scale alignment.
%
Feat. stands for ``\textit{Features}'', BA stands for ``\textit{Balanced Accuracy}", and A stands for ``\textit{Accuracy}''.}
\label{table1}
 \end{minipage}
 \vspace{-1em}
\end{table*}

\subsection{Training Strategy}
\textbf{Stage 1: Hierarchical Cross-Modal Alignment:} Recent SOTA CPath models align global WSI embeddings with entire pathology reports \cite{li2024llava, chen2025slidechat}, which limits fine-grained semantic alignment and degrades VQA performance (Fig.~\ref{fig2}). 
MLLM-HWSI achieves hierarchical visual–textual alignment across multiple levels via \textit{hierarchical contrastive and cross-scale consistency objectives}, capturing the linguistic hierarchy of pathology reports. 
This stage utilizes 9,642 WSI–report pairs \cite{liang2025wsillavamultimodallargelanguage}, updating all hierarchical encoders ($\textrm{ViT}{\textrm{cell-cell}}$, $\mathcal{F}{\textrm{CONCH}}$, $\textrm{ViT}{r}$, $\textrm{ViT}{\textrm{WSI}}$) and the text encoder, while keeping the VL projectors and LLM weights frozen.
Let the token embeddings of a pathology report be $\mathbf{T} = \{ t_1, t_2, \ldots, t_M \}$, the scale-specific contrastive loss is:
\vspace{-3mm}
\begin{equation}
\mathcal{L}_{s} = -\frac{1}{n_s}\sum_i \log
\frac{\exp(\text{sim}(z_{s,i}, t_{i}) / \tau)}
{\sum_j \exp(\text{sim}(z_{s,i}, t_{j}) / \tau)},
\vspace{-3mm}
\end{equation}
where $s \in \{c, p, r\}$ represents the cell-, patch-, and region-level, and $n_s$ denotes the number of visual tokens at that level, and $\tau$ is a temperature parameter controlling distribution sharpness. 
Each $t_j$ corresponds to the $j^{\text{th}}$ token embedding from the pathology report, serving as a contrastive negative in the denominator.
At the WSI-level, we use an analogous formulation:
\vspace{-3mm}
\begin{equation}
\mathcal{L}_{\textrm{WSI}}= -\frac{1}{n_b}\sum_b \log
\frac{\exp(\text{sim}(z_{\textrm{WSI},b}, t_{r,b}) / \tau)}
{\sum_{l=1}^{n_{b}} \exp(\text{sim}(z_{\textrm{WSI},b}, t_{r,l}) / \tau)},
\label{eqn13}
\vspace{-3mm}
\end{equation}
where $n_b$ denotes the batch size, and $t_{r,b}$ represents the textual embedding of the pathology report associated with each WSI.
While $\mathcal{L}_{s}$ and $\mathcal{L}_{\textrm{WSI}}$ ensure local alignment at individual scales, they do not ensure semantic consistency between adjacent levels (e.g., patch vs.\ region, region vs.\ WSI) leading to semantic drift across scales. 
To address this issue, we introduce a \textbf{cross-scale consistency loss} that promotes hierarchical coherence by encouraging smooth transitions from fine- to coarse-grained representations:
\vspace{-3mm}
\begin{equation}
\begin{split}
   \mathcal{L}_{c} =  \frac{1}{2n_{r}}\sum_{s \in \{c, p\}} \sum_{k=1}^{n_{r}} \big\| z_{r,k} - \frac{1}{n_{s}}\sum_{i=1}^{n_{s}} z_{s,k,i} \big\|_2^2 
   \\
   + \frac{1}{n_{p}}\sum_{j=1}^{n_{p}} \big\| z_{c_{j}} - z_{p_{j}} \big\|_2^2.
\end{split}
\label{eqn14}
\vspace{-4mm}
\end{equation}
\noindent The total hierarchical alignment loss, denoted as $\mathcal{L}_{\textrm{HCA}}$, integrates all scale-specific objectives as: 
\vspace{-2mm}
\begin{equation}
\mathcal{L}_{\textrm{HCA}} = \frac{1}{n_{b}}\sum_{k=1}^{n_{b}} 
(\mathcal{L}^{k}_{s \in \{c,p,r\}} + \mathcal{L}^{k}_{c}) 
+ \mathcal{L}_{\textrm{WSI}}.
\label{eqn15}
\vspace{-3mm}
\end{equation}
This stage ensures semantically consistent hierarchical visual-pathology report alignment.

\noindent \textbf{Stage 2: Feature Space Alignment:} In this stage, the pretrained hierarchical encoders are combined with the V–L projectors and the LLM. 
Only the projection matrices are trained on 9,642 WSI–report pairs \cite{liang2025wsillavamultimodallargelanguage}. 

\noindent \textbf{Stage 3: Task-Specific Instruction Tuning:}
In this stage, the projection matrices and LLM are jointly fine-tuned using 175,450 WSI-level VQA pairs \cite{liang2025wsillavamultimodallargelanguage}. 
This stage enables the model to perform task-specific reasoning, including WSI-level diagnostic classification, report generation, and VQA, by leveraging the aligned multi-scale visual–textual representations learned in previous stages.
\section{Experiments}
\label{sec:results}
\textbf{Training and Implementation:} Stage~1 pretraining uses 9{,}642 WSI–caption (report) pairs \cite{liang2025wsillavamultimodallargelanguage}, and train for 50 epochs with learning rate $10^{-3}$, $n_{b}$ $64$, and $\tau$ $0.02$.
All encoders including $\textrm{ViT}_{\textrm{cell}\textrm{-}\textrm{cell}}$, $\mathcal{F}_{\textrm{CONCH}}$, $\textrm{ViT}_{r}$, and $\textrm{ViT}_{\textrm{WSI}}$, and the text encoder are fine-tuned. 
$\textrm{ViT}_{\textrm{cell}\textrm{-}\textrm{cell}}$ contains two transformer blocks with two self-attention heads.
We employed Qwen2.5-7B-Instruct as backbone LLM \cite{yang2024qwen2} during pretraining.
In Stage 2, two-layer hierarchical VL projectors are trained with batch size $256$.  
In Stage 3, we used WSI-Bench \cite{liang2025wsillavamultimodallargelanguage} with learning rate $2\times10^{-5}$, and batch size $128$. 
We adopt LoRA (rank $128$, $\alpha=256$) and leverage DeepSpeed ZeRO-3 for distributed training. 
All experiments are run on 4 NVIDIA A100 80GB GPUs.

\begin{table*}[t]
\begin{minipage}{0.72\textwidth}
   \centering \small
  \setlength{\tabcolsep}{5.0pt}
\centering
\scalebox{1.0}{
\begin{tabular}{l|c c c|c cccc}
\hline
\multirow{2}{*}{Variants} &  \multicolumn{3}{c|}{$\mathcal{L}_{HCA}$ Loss (\ref{eqn15})} &PANDA &EBRAINS&WSI-VQA& SlideBench-VQA \\
 &$\mathcal{L}_{s}$&$\mathcal{L}_{c}$&$\mathcal{L}_{\textrm{WSI}}$&(BA)&(BA)&(A)&(BCNB) (A)\\
\hline
MLLM-HWSI   & $\checkmark$ &  $\checkmark$&$\checkmark$ &\textbf{0.748}&\textbf{0.612}&\textbf{0.692}&\textbf{0.687}\\
MLLM-HWSI   & $\checkmark$ &  $\checkmark$  &$\times$ &0.716&0.592&0.668&\underline{0.654}\\
MLLM-HWSI   & $\checkmark$ &  $\times$  &$\times$ &0.705&0.582&0.655&0.636\\
MLLM-HWSI   & $\times$ &  $\times$&$\checkmark$&0.661&0.519&0.616&0.576\\
MLLM-HWSI   & $\times$ &  $\checkmark$&$\checkmark$ &0.677&0.526&0.624&0.590\\
MLLM-HWSI   & $\checkmark$ &  $\times$&$\checkmark$ &\underline{0.723}&\underline{0.605}&\underline{0.671}&0.653\\
\hline
\end{tabular}
}
\end{minipage}
  \hfill
  \begin{minipage}{0.28\textwidth}
  \small \vspace{1em}
\caption{
\textbf{Ablation 7: Effect of loss components in $\mathcal{L}_{HCA}$.}
Removing any of the semantic ($\mathcal{L}_{s}$), cross-scale ($\mathcal{L}_{c}$), or WSI-level ($\mathcal{L}_{WSI}$) losses leads to notable performance drops, confirming their complementary contributions to hierarchical cross-modal alignment.
}
\label{table7}
 \end{minipage}
 \vspace{-1em}
\end{table*}

 \begin{figure*}[t!]
\centering 
 \begin{subfigure}[t]{0.331\textwidth} 
 \centering 
 \includegraphics[width=\textwidth, height=3.2cm]{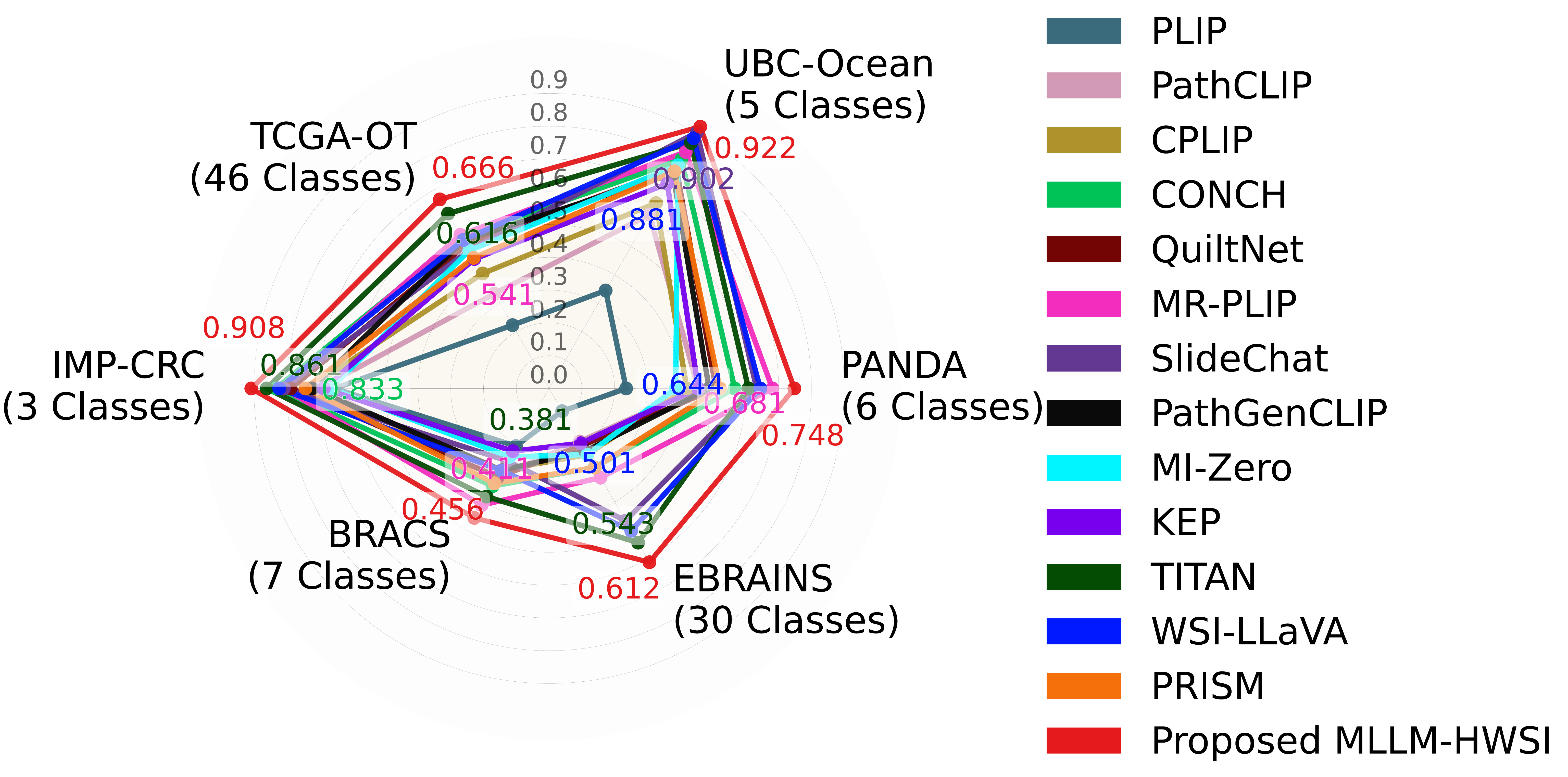} 
 \caption{Zero-shot WSI Classification Performance} 
 \label{fig:patch_a} 
 \end{subfigure}%
 \hfill 
 \begin{subfigure}[t]{0.331\textwidth} 
 \centering 
 \includegraphics[width=\textwidth, height=3.2cm]{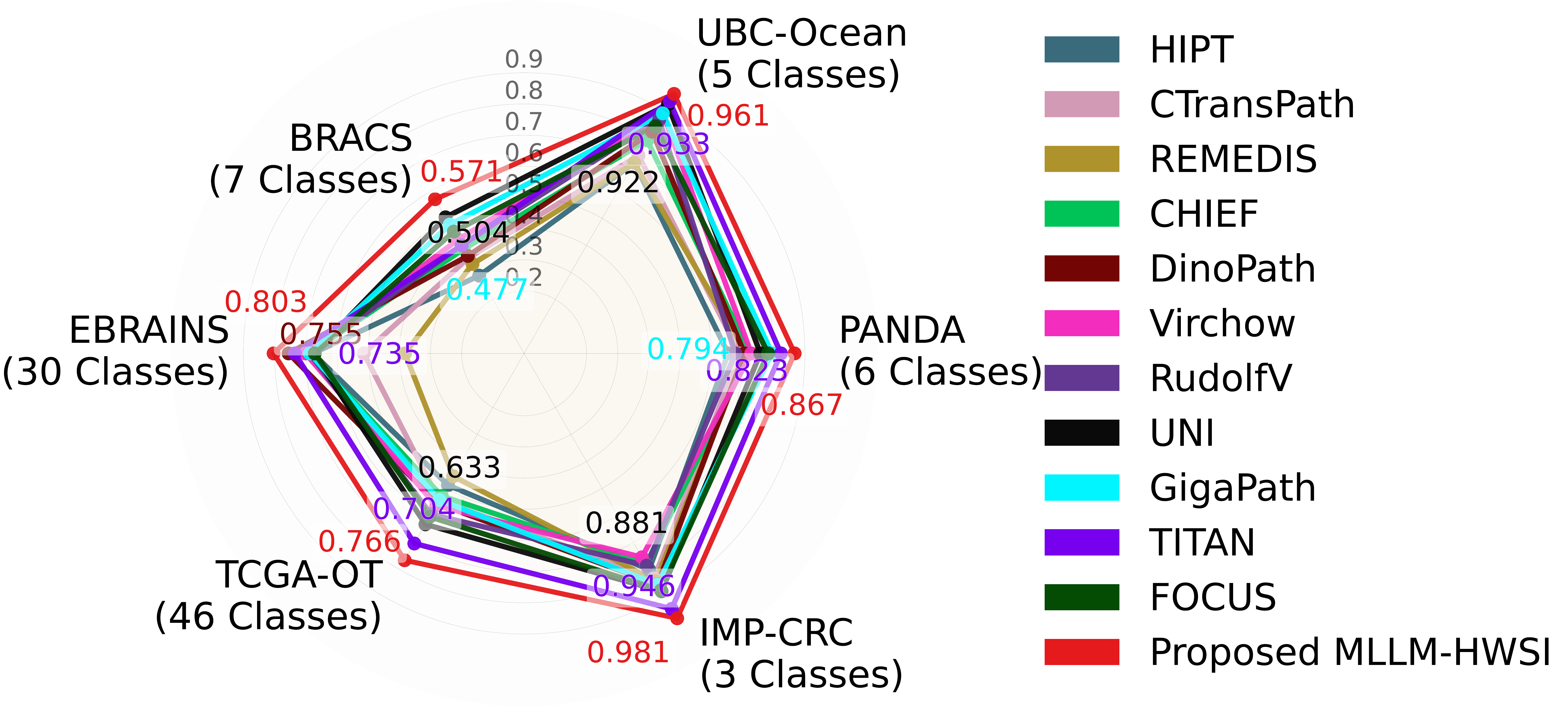}
 \caption{WSI Classification Using Linear Probing} 
 \label{fig:patch_b} 
 \end{subfigure} 
  \hfill 
 \begin{subfigure}[t]{0.331\textwidth} 
 \centering 
 \includegraphics[width=\textwidth, height=3.2cm]{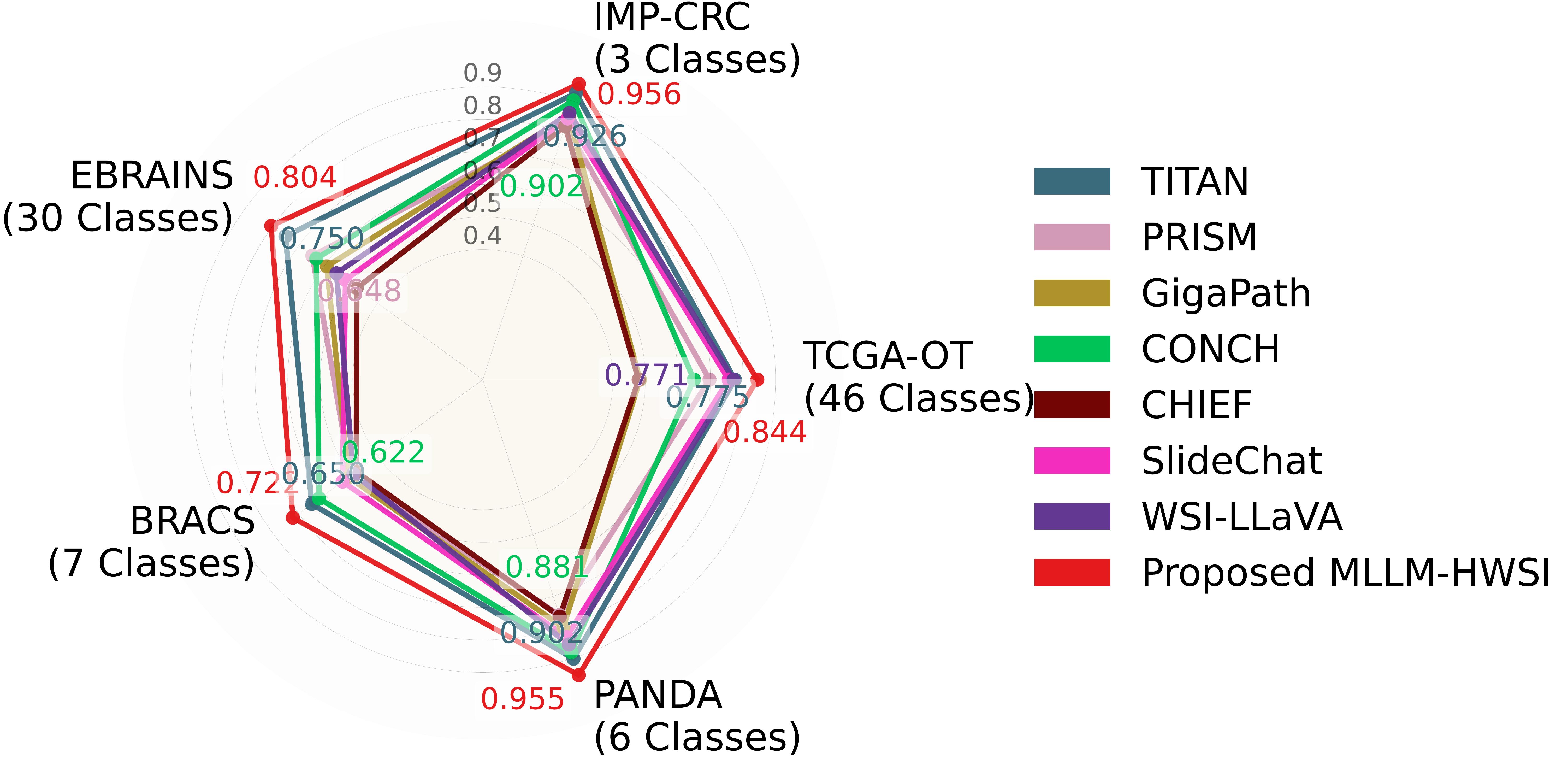}
   \caption{Zero-shot WSI Retrieval} 
 \label{fig:patch_c} 
 \end{subfigure} 
 \vspace{-3mm} 
\caption{
\textbf{Performance comparison of the proposed MLLM-HWSI with SOTA CPath models.}
MLLM-HWSI achieves the highest overall scores across all benchmarks, underscoring the benefits of hierarchical multi-scale visual encoding and cross-modal alignment.
}
 \label{fig:radar}
 \vspace{-4mm}
 \end{figure*} 

\noindent \textbf{CPath Tasks and Datasets:} MLLM-HWSI is evaluated on six WSI-level tasks.
For classification (zero-shot and linear probe), we use BRACS (7 classes) \cite{brancati2022bracs}, UBC-Ocean (5) \cite{UBC-OCEAN}, TCGA-OT (46) \cite{ding2024titan,mahmood2024tcgaot}, EBRAINS (30) \cite{roetzer2022digital}, PANDA (6) \cite{Bulten2022_PANDA}, and IMP-CRC (3) datasets \cite{neto2024interpretable}. 
Zero-shot VQA is assessed on WSI-Bench (4{,}119 pairs) \cite{Liang2024_WSI-LLaVA}, WSI-VQA (8{,}672) \cite{chen2025wsi}, SlideBench-VQA (BCNB: 7{,}247) \cite{chen2025slidechat}, and SlideBench-VQA (TCGA: 7{,}824) \cite{chen2025slidechat}. 
Report generation is evaluated on WSI-Bench (208 WSI–report pairs) \cite{liang2025wsillavamultimodallargelanguage} and HistGen (700) \cite{guo2024histgen}. 
WSI retrieval uses TCGA-OT \cite{ding2024titan,mahmood2024tcgaot}, EBRAINS \cite{roetzer2022digital}, and IMP-CRC \cite{neto2024interpretable}. 
Cross-modal retrieval is measured on TCGA Reports \cite{weinstein2013cancer,ding2024titan}, and caption generation on SlideBench \cite{chen2025slidechat}.


\subsection{Evaluation Metrics and SOTA Comparisons}
For classification, we employed weighted $F_{1}$ and Balanced Accuracy (BA) \cite{chen2024towards}, for report/caption generation, ROUGE, BLEU-1–4, and METEOR \cite{chen2025slidechat,seyfioglu2024quilt}, for VQA accuracy \cite{seyfioglu2024quilt}, for cross-modal retrieval Recall@K \cite{ding2024titan}, for WSI retrieval Top-1\% accuracy \cite{ding2024titan}.

For zero-shot classification and WSI retrieval, we compare against 10 SOTA CPath VLMs: PLIP \cite{huang2023visual}, PathCLIP \cite{sun2024pathasst}, MI-Zero \cite{lu2023visual}, CONCH \cite{lu2024visual}, QuiltNet \cite{ikezogwo2024quilt}, CPLIP \cite{javed2024cplip}, MR-PLIP \cite{albastaki2025multi}, PathGenCLIP \cite{sun2024pathgen}, TITAN \cite{ding2024titan}, KEP \cite{zhou2024knowledge}, and PRISM \cite{shaikovski2024prism}. 
We use dataset-specific prompts as recommended by CONCH \cite{lu2024visual}. 
For linear-probe and weakly supervised settings, we compare with HIPT \cite{chen2022scaling}, TITAN \cite{ding2024titan}, UNI \cite{chen2024towards}, CTransPath \cite{wang2022transformer}, REMEDIS \cite{azizi2022robust}, CHIEF \cite{wang2024pathology}, DINOPath \cite{kang2023benchmarking}, Virchow \cite{vorontsov2024foundation}, GigaPath \cite{xu2024whole}, and RudolfV \cite{dippel2024rudolfv}.
For VQA, report/caption generation, and cross-modal retrieval, we benchmark against general LMMs—GPT-4V \cite{hurst2024gpt}, LLaVA \cite{liu2024improved}, Qwen-VL-Max \cite{bai2023qwenvlversatilevisionlanguagemodel}, and Gemini-Pro-Vision \cite{team2023gemini}, as well as CPath-specific MLLMs: Quilt-LLaVA \cite{seyfioglu2024quilt}, SlideChat \cite{chen2025slidechat}, WSI-LLaVA \cite{liang2025wsillavamultimodallargelanguage}, PRISM \cite{shaikovski2024prism}, MedDr \cite{he2024meddr}, LLaVA-Med \cite{li2023llava}, HistGen \cite{guo2024histgen}, and PathGen-LLaVA \cite{sun2024pathgen}. 
For fairness, we use official code, consistent test splits, and identical inference prompts. 

\subsection{Ablation Studies}
\noindent\textbf{1.\ Importance of Hierarchical Representations:}
As shown in Table~\ref{table1}, we progressively augment the hierarchical features in MLLM-HWSI$_{1-3}$. 
Using only WSI-level features (MLLM-HWSI$_1$) already exceeds baseline methods. 
Adding region, patch, and cell-level features yields consistent improvements across all datasets. 
A complementary \emph{subtractive} study (MLLM-HWSI$_{4-7}$) causes notable drops, underscoring the importance of every representation level. \textbf{2.\ Loss Function:} Table~\ref{table7} further analyzes the hierarchical cross-modal alignment loss $\mathcal{L}_{\textrm{HCA}}$ (Eq.~\ref{eqn15}) by removing each term in turn. 
Dropping any component degrades performance. 
With only the WSI-level loss $\mathcal{L}_{\textrm{WSI}}$ (Eq.~\ref{eqn13}) retained (i.e., removing both $\mathcal{L}_{c}$ and $\mathcal{L}_{s}$), WSI-level classification declines by 8.70\% (PANDA) and 9.30\% (EBRAINS), while VQA accuracy drops by 7.60\% (WSI-VQA) and 11.10\% (SlideBench). 
These results confirm the necessity of cross-modal alignment across hierarchices.

\begin{table*}[t]
\begin{minipage}{0.72\textwidth}
   \centering \small
  \setlength{\tabcolsep}{1.6pt}
\centering
\scalebox{0.9}{
\begin{tabular}{l|c c cc|cccc|cc}
\hline
\multirow{2}{*}{General MLLMs}&\multicolumn{4}{c|}{SlideBench-VQA (TCGA)} & \multicolumn{4}{c|}{WSI-Bench}&SlideBench-&\multirow{2}{*}{WSI-VQA}\\
& Micro. & Diag. &  Clinical &Average&MA&Diag.&TP&Average&VQA(BCNB) &\\
\hline
InstructBLIP-FLAN & 0.366 &0.186&0.221&0.257&0.198&0.221&0.389&0.269&0.189&0.102\\
LLaVA-1.5& 0.451 & 0.219 & 0.389&0.353&0.232&0.271&0.677&0.393&0.201&0.121\\
Qwen-VL-MAX &0.496&0.288&0.405&0.396&0.288&0.322&0.706&0.438&0223&0.133\\
GeminiProV &0.506&0.304&0.587&0.465&0.403&0.433&0.821&0.552&0.282&0.167\\
GPT-4V&0.628&0.466&0.667&0.587&0.471&0.530&0.875&0.625&0.414&0.304\\
\hline
\multirow{2}{*}{CPath MLLMs}&\multicolumn{4}{c|}{SlideBench-VQA (TCGA)} & \multicolumn{4}{c|}{WSI-Bench}&SlideBench-&\multirow{2}{*}{WSI-VQA}\\
& Micro. & Diag. &  Clinical &Average&MA&Diag.&TP&Average&VQA(BCNB) &\\
\hline
LLaVA-Med&0.458&0.275&0.408&0.803&0.866&0.732&0.912&0.836&0.124&0.187\\
Quilt-LLaVA &0.491& 0.269& 0.447& 0.402&0.947&0.849&\textbf{1.000}&0.932&0.415&0.354\\
PathGen-LLaVA & 0.566 & 0.321 & 0.509&0.465&0.882&0.781&0.922&0.861&0.401&0.331\\
MedDr &0.733&0.577&0.742&0.684&0.902&0.831&0.922&0.885&0.336&0.543\\
WSI-VQA & 0.334 & 0.189 &0.306&0.276&0.758&0.577&0.771&0.702&0.113&0.469\\
TITAN &0.851&0.745&0.824&0.806&0.940&\underline{0.883}&\textbf{1.000}&\textbf{0.941}&0.551&0.586\\
SlideChat &0.876&0.732&\underline{0.842}&0.816&0.932&0.858&0.971&0.920&0.541&\underline{0.601}\\
WSI-LLaVA &\underline{0.882}&\underline{0.752}&0.841&\underline{0.825}&\underline{0.951}&0.863&\textbf{1.000}&0.938&\underline{0.553}&0.546\\
MLLM-HWSI &\textbf{0.956} &\textbf{ 0.824} & \textbf{0.908}&\textbf{0.896}&\textbf{0.989}&\textbf{0.962}&\underline{0.986}&\textbf{0.979}&\textbf{0.687}&\textbf{0.692}\\
\hline
\end{tabular}
}
\end{minipage}
  \hfill
  \begin{minipage}{0.30\textwidth}
  \small \vspace{1em}
\caption{
\textbf{Comparison of MLLM-HWSI with SOTA general-purpose and CPath-specific MLLMs on multi-domain VQA benchmarks.}
We evaluate MLLM-HWSI across four datasets, two external (SlideBench-VQA (BCNB), WSI-VQA) and two TCGA-based (SlideBench-VQA (TCGA), WSI-Bench), covering Microscopy \textit{Micro}, Diagnosis \textit{Diag.}, Morphological Analysis \textit{MA}, and Treatment Panning \textit{TP}–related questions. 
Performance is reported in terms of accuracy. 
MLLM-HWSI achieves superior accuracy across all datasets and sub-tasks, demonstrating its strong generalization and diagnostic reasoning capabilities.
}
\label{table9}
 \end{minipage}
 \vspace{-4mm}
\end{table*}

\begin{table*}[t!]
\begin{minipage}{0.72\textwidth}
   \centering \small
  \setlength{\tabcolsep}{3.5pt}
\centering
\scalebox{0.98}{
\begin{tabular}{l|cccccc}
\toprule
Models& BLEU-1& BLEU-2&BLEU-3& BLEU-4&ROUGE&METEOR\\
\hline
GPT-4V& 0.202$|$0.158 & 0.069$|$0.420 & 0.030$|$0.240 & 0.016$|$0.100 & 0.132$|$0.128 & 0.167$|$0.144\\
Quilt-LLaVA & 0.474$|$0.384 & 0.351$|$0.364 & 0.282$|$0.301 & 0.236$|$0.254 & 0.475$|$0.441 & 0.460$|$0.462\\
MIGen&0.403$|$0.402&0.306$|$0.466&0.248$|$0.266 & 0.209$|$0.234&0.446$|$0.322 & 0.407$|$0.412  \\
WSICaption&0.203$|$0.218&0.156$|$0.137&0.183$|$0.460&0.130$|$0.540&0.265$|$0.251&0.317$|$0.322 \\
TITAN&0.399$|$0.382&0.321$|$0.318&0.223$|$0.247&0.203$|$0.314&0.304$|$0.341&0.385$|$0.329 \\
HistGen& 0.406$|$0.413 & 0.307$|$0.297 & 0.248$|$0.229 & 0.208$|$ 0.184& 0.448$|$0.344 &0.416$|$0.182 \\
SlideChat&0.441$|$0.551&0.310$|$\underline{0.533}&0.277$|$0.379&0.191$|$\underline{0.655}&0.463$|$0.492&0.422$|$0.613  \\
WSI-LLaVA&\underline{0.480}$|$\underline{0.592}&\underline{0.358}$|$0.527&\underline{0.287}$|$\underline{0.488}&\underline{0.240}$|$0.644&\underline{0.490}$|$\underline{0.544}&\underline{0.465}$|$\underline{0.655}\\
MLLM-HWSI&\textbf{0.556}$|$\textbf{0.667}&\textbf{0.426}$|$\textbf{0.570}&\textbf{0.348}$|$\textbf{0.566}&\textbf{0.287}$|$\textbf{0.685}&\textbf{0.551}$|$\textbf{0.605}&\textbf{0.513}$|$\textbf{0.706} \\
\hline
\end{tabular}
}
\end{minipage}
  \hfill
  \begin{minipage}{0.30\textwidth}
  \small \vspace{1em}
\caption{
\textbf{Report generation comparison on two benchmarks WSI-Bench $|$ HistGen.}  
MLLM-HWSI outperforms all SOTA models across BLEU, ROUGE-L, and METEOR metrics, highlighting its ability to produce accurate and clinically coherent diagnostic reports.
}
\label{table10}
 \end{minipage}
 \vspace{-1em}
\end{table*}

\begin{table}[t]
\centering
\resizebox{\columnwidth}{!}{
\begin{tabular}{l|ccccc}
\toprule
Models&R$@$1& R$@$3& R$@$5 & R$@$10 &Average\\
\hline
PRISM & 0.542$|$0.634   &        0.674$|$0.768       &        0.724$|$0.817  &         0.767$|$0.880      &        0.669$|$0.765\\
TITAN & 0.732$|$0.754 &         0.818$|$\underline{0.896} &       0.841$|$\underline{0.908}      &          \underline{0.876}$|$\underline{0.922} &   0.811$|$\underline{0.866}\\
SlideChat &0.766$|$\underline{0.771}  &    0.823$|$0.855       &          0.838$|$0.884   &          0.852$|$0.907       &        0.819$|$0.854\\
WSI-LLaVA &       \underline{0.786}$|$0.770       &           \underline{0.824}$|$0.867          &           \underline{0.844}$|$0.890         &           0.858$|$0.912        &      \underline{0.828} $|$0.859\\
MLLM-HWSI &       \textbf{0.822}$|$\textbf{0.842}          &            \textbf{0.856}$|$\textbf{0.942}      &            \textbf{0.897}$|$\textbf{0.952}        &           \textbf{0.926}$|$\textbf{0.946}        &       \textbf{0.875}$|$\textbf{0.920}\\
\hline
\end{tabular}
}
\vspace{-3mm}
\caption{
\textbf{Cross-modal retrieval on TCGA-Slide-Reports.}  
MLLM-HWSI consistently outperforms SOTA models in both report-to-slide and slide-to-report tasks across all recall metrics.
}
\label{table11}
\vspace{-4mm}
\end{table}

\begin{table}[t!]
\centering
\resizebox{\linewidth}{!}{
\begin{tabular}{l|cccccc}
\toprule
Models& BLEU-1& BLEU-2&BLEU-3& BLEU-4&ROUGE-L&METEOR \\
\hline
GPT-4V& 0.100 & 0.030 & 0.010 & 0.010 & 0.110 & 0.131 \\
Quilt-LLaVA& 0.230 & 0.090 & 0.040 & 0.010 & 0.160 & 0.420 \\
MIGen & 0.370 & 0.240 & 0.150 & 0.100 & 0.250 & 0.381 \\
HistGen&0.300&0.181&0.110&0.090&0.171&0.288 \\
SlideChat & 0.370 & 0.210 & 0.120 & 0.080 & 0.240 & 0.488\\
TITAN & 0.377& 0.233& 0.169 & \underline{0.176} & 0.302& 0.506\\
PRISM &0.302& 0.168& 0.156 & 0.161 & 0.266& 0.423\\
WSI-LLaVA&\underline{0.411}&\underline{0.266}&\underline{0.180}&0.150&\underline{0.320}&\underline{0.551}\\
MLLM-HWSI&\textbf{0.462}&\textbf{0.324}&\textbf{0.267}&\textbf{0.231}&\textbf{0.367}&\textbf{0.627} \\
\hline
\end{tabular}
}
\vspace{-3mm}
\caption{
\textbf{Captioning performance on SlideBench-Caption.}  
MLLM-HWSI outperforms all SOTA models showing strong capability in producing morphology-aware and accurate captions.
}
\label{table12}
\vspace{-5mm}
\end{table}

\subsection{Main Results}
\textbf{1. Zero-shot WSI Classification:} The proposed MLLM-HWSI is compared against 13 SOTA CPath VLMs and MLLMs across five external and one internal dataset in terms of BA (Fig. \ref{fig:radar} (a)).  
\textit{MLLM-HWSI model achieves an average BA of 71.86\%, surpassing TITAN (64.56\%) and WSI-LLaVA (61.01\%) by 7.30\% and 10.85\%, respectively. }
This consistent improvement highlights the effectiveness of hierarchical multi-scale alignment. \textbf{2.\ Linear Probe Evaluation:} We compare MLLM-HWSI with 11 SOTA vision-only and VL CPath models across six datasets using linear probe and weakly supervised classification in terms of BA (Fig.~\ref{fig:radar}(b)).  
\textit{MLLM-HWSI achieves an average BA of 82.48\%, outperforming TITAN (75.68\%) and UNI (72.86\%) by 6.80\% and 9.62\%, respectively.} These results emphasize the contribution of our hierarchical multi-scale visual representations to more discriminative feature learning. \textbf{3. WSI Retrieval:} MLLM-HWSI is evaluated on five datasets for zero-shot retrieval performance using top-1\% accuracy (Fig.~\ref{fig:radar}(c)).  
\textit{MLLM-HWSI achieves an average performance of 85.62\%, outperforming TITAN (80.06\%) and CONCH (73.74\%) by 5.56\% and 11.88\%, respectively.} These improvements validate the benefit of hierarchical multi-scale representation alignment for accurate WSI retrieval.
\noindent \textbf{4. WSI VQA:} We evaluate MLLM-HWSI on four VQA benchmarks to assess multi-scale reasoning and diagnostic comprehension across morphological, clinical, and pathological tasks (Table~\ref{table9}). 
These benchmarks require both detailed cell-level analysis and holistic WSI-level interpretation, offering a rigorous test of multimodal reasoning.
MLLM-HWSI consistently outperforms both general-purpose MLLMs and pathology-specific models.  
\textit{On average, it achieves 89.60\% accuracy on SlideBench-VQA (TCGA), 68.70\% on SlideBench-VQA (BCNB), 97.90\% on WSI-Bench, and 69.20\% on WSI-VQA—surpassing all previous SOTA results.} These gains stem from MLLM-HWSI’s hierarchical visual representations, cross-scale VL alignment via consistency-regularized loss, and instruction fine-tuning that strengthens context-aware clinical reasoning. 
\noindent \textbf{5. WSI Report Generation:} We evaluate MLLM-HWSI for report generation using WSI-Bench and HisGen datasets, comparing against both general-purpose and CPath-specific models (Table~\ref{table10}).  
\textit{MLLM-HWSI achieves the best performance across all metrics.} These results surpass all prior SOTA models, \textit{demonstrating MLLM-HWSI’s ability to generate accurate, clinically coherent, and morphology-aware diagnostic reports.} The performance gains arise from \textit{hierarchical visual alignment and cross-scale consistency that capture both fine-grained morphology and high-level diagnostic context}. \noindent \textbf{6. WSI Caption Generation:} For caption generation on the SlideBench-Caption dataset, MLLM-HWSI achieves the best results across all metrics (Table \ref{table10}).
\textit{It achieves BLEU-1/2/3/4 of 46.20\%, 32.40\%, 26.70\%, and 23.10\%, with ROUGE-L = 36.70\% and METEOR = 62.70\%, surpassing WSI-LLaVA by a notable margin. } 
These results highlight the model’s strong ability to produce concise, morphology-aware, and clinically relevant captions that faithfully summarize WSI-level findings.
\noindent \textbf{7. WSI Cross-Modal Retrieval:} We evaluate cross-modal retrieval performance using Recall@K metrics.  
\textit{MLLM-HWSI achieves consistent gains, outperforming WSI-LLaVA by 4.70\% and 6.10\% on both tasks (Table \ref{table11}).}
\textit{These findings validate MLLM-HWSI’s strong alignment between textual and hierarchical visual modalities, enabling accurate and interpretable retrieval through consistency-regularized hierarchical VL alignment.}

\section{Conclusion}
\label{sec:conclusion}
\vspace{-1em}
We presented a hierarchical multimodal LLM in CPath that leverages multi-scale VL alignment across WSI to enhance diagnostic understanding in key tasks such as VQA, captioning, and report generation. 
It decomposes WSIs into a hierarchical representation comprising cell, patch, region, and WSI-level embeddings. 
Each hierarchy is aligned with textual semantics via dedicated VL projectors integrated into a MLLM, enabling multi-granular reasoning across spatial scales.
The proposed optimization objective combines three complementary components including cross-modal alignment, hierarchical feature-space consistency,  and instruction fine-tuning to enhance diagnostic reasoning. 
Comprehensive experiments across six CPath tasks demonstrate that MLLM-HWSI consistently surpasses SOTA models, validating the effectiveness of hierarchical multi-scale alignment and cross-modal reasoning.  
By unifying hierarchical visual understanding with language-driven inference, MLLM-HWSI establishes a new paradigm for interpretable foundation models in CPath, offering potential to assist expert pathologists in clinical decision-making.  
In future work, we aim to extend MLLM-HWSI beyond histopathology toward broader multimodal medical integration including radiology, genomics, and clinical records—to enable holistic, patient-level reasoning within a unified medical AI framework.

\section{Acknowledgement} 
This research was funded by Khalifa University of Science and Technology through the Faculty Start-Ups under the grant number: KU-INT-FSU-2005-8474000775.

{
    \small
    \bibliographystyle{ieeenat_fullname}
    \bibliography{main}
}

\newpage

\twocolumn[
\begin{center}
	\Huge \textbf{Supplementary Material}\\MLLM-HWSI: A Multimodal Large Language Model for Hierarchical Whole Slide Image Understanding
\end{center}
\vspace{1em}
]

\section{Inference Details}
Each WSI is partitioned into $\approx$20 regions, each with 256 patches.
Since SPF has two components: (i) HPS (Eq. 1), which removes redundant patches using visual similarity only, and (ii) DPS (Eq. 2), which leverages report-derived semantic tokens to guide patch relevance during training.
Therefore, during inference, no pathology reports are used.
Only HPS is applied, so patch selection is fully vision-based with no test-time information leakage.

\section{Hierarchical WSI-Caption Alignment}
\label{alignment}
In Computational Pathology (CPath), the importance of hierarchical alignment arises from both biological reasoning and representational learning principles \cite{cui2021artificial, song2023artificial, fuchs2011computational}.
Theoretically, WSIs are not uniform visual entities; instead, they exhibit a nested organization, where meaning emerges across multiple levels of abstraction \cite{cornish2012whole, chen2022scaling, zarella2019practical, hanna2020whole}. 
Diagnostic semantics are inherently hierarchical: cellular morphology defines nuclear atypia and mitotic figures; patch-level structures capture gland formation, necrosis, or immune infiltration; region-level context reflects tumor invasion and stromal interaction; and the global WSI conveys architectural disarray and overall differentiation \cite{kiran2023digital, betmouni2021diagnostic, dimitriou2024magnifying}.
A single global embedding, as used in conventional MLLMs \cite{chen2025slidechat, liang2025wsillavamultimodallargelanguage}, collapses this structure and causes information loss, particularly of the spatial and semantic dependencies that exist between local and global tissue organization. 
Hierarchical alignment mitigates this by learning distinct yet interconnected visual–language mappings for each scale.
Each level aligns with its corresponding linguistic abstraction—cells correspond to morphological words, patches to descriptive phrases, regions to structural sentences, and the WSI to a diagnostic paragraph—thus preserving compositional semantics and ensuring that information propagates coherently across scales \cite{hutter2018cancer, fraggetta2017routine, brixtel2022whole}.

\begin{table*}[h!]
\centering
\resizebox{\linewidth}{!}{%
\begin{tabular}{l|l|llcc|}
\hline
Cell Segmentation&Cell&PANDA& EBRAINS &WSI-VQA& SlideBench-VQA\\
Backbone&Feat.&\cite{Bulten2022_PANDA} (BA)&\cite{roetzer2022digital} (BA)&\cite{chen2025wsi} (A)&(BCNB) \cite{chen2025slidechat}(A)\\
\hline
CellViT \cite{horst2024cellvit}&$\textrm{ViT}_{\textrm{cell-cell}}$&\textbf{0.748}&\textbf{0.612}&\textbf{0.692}&\textbf{0.687}\\
NuHTC \cite{li2025nuhtc}& $\textrm{ViT}_{\textrm{cell-cell}}$&\underline{0.733}&\underline{0.600}&0.689&\underline{0.685}\\
STRARDIST \cite{weigert2022} &$\textrm{ViT}_{\textrm{cell-cell}}$&0.701&0.585&0.665&0.670\\
MicroNet \cite{RAZA2019160}&$\textrm{ViT}_{\textrm{cell-cell}}$ &0.698&0.561&0.672&0.671\\
HoverNet \cite{graham2019hover}& $\textrm{ViT}_{\textrm{cell-cell}}$&0.725&0.591&\underline{0.690}&0.673\\
\hline
\end{tabular}
}
\caption{
\textbf{Effect of cell segmentation backbones in $\textrm{ViT}_{\textrm{cell-cell}}$.}
Results show Balanced Accuracy (BA) for PANDA and EBRAINS, and Accuracy (A) for WSI-VQA and SlideBench-VQA (BCNB). 
CellViT achieves the highest scores, confirming the benefit of SAM-based segmentation for cell-level feature extraction.
}
\label{table1}
\end{table*}

\begin{table*}[h!]
\centering
\resizebox{\linewidth}{!}{%
\begin{tabular}{l|cccc|cccc}
\hline
\multirow{2}{*}{Models}&Cell&Patch&Region&WSI&PANDA& EBRAINS &WSI-VQA& SlideBench-VQA\\
&Encoder&Encoder&Encoder&Encoder&\cite{Bulten2022_PANDA} (BA)&\cite{roetzer2022digital} (BA)&\cite{chen2025wsi} (A)&(BCNB) \cite{chen2025slidechat} (A)\\
\hline
MLLM-HWSI&$\textrm{ViT}_{\textrm{cell-cell}}$ &$\mathcal{F}_{\textrm{CONCH}}$&$\textrm{ViT}_{r}$&$\textrm{ViT}_{\textrm{WSI}}$&\textbf{0.748}&\textbf{0.612}&\textbf{0.692}&\textbf{0.687}\\\
MLLM-HWSI&   $\textrm{ViT}_{\textrm{cell-cell}}$ &UNI &UNI&UNI&0.721&\underline{0.589}&0.653&0.664\\
MLLM-HWSI&$\textrm{ViT}_{\textrm{cell-cell}}$ &CONCH &CONCH&CONCH&\underline{0.712}&0.581&0.644&0.657\\
MLLM-HWSI& $\textrm{ViT}_{\textrm{cell-cell}}$&CONCH &CONCH&LongNet&0.702&0.575&\underline{0.681}&0.665\\
MLLM-HWSI& $\textrm{ViT}_{\textrm{cell-cell}}$&GigaPath &GigaPath&LongNet&0.692&0.562&0.673&\underline{0.681}\\
MLLM-HWSI&$\textrm{ViT}_{\textrm{cell-cell}}$ &UNI &UNI&LongNet&0.686&0.564&0.663&0.634\\
\hline
\end{tabular}
}
\caption{
\textbf{Influence of visual encoder selection across hierarchical levels.}
Different combinations of patch-, region-, and WSI-level encoders (UNI, CONCH, GigaPath, LongNet) are evaluated, all fine-tuned with the proposed loss. 
The $\mathcal{F}_{\textrm{CONCH}}$, $\textrm{ViT}_{r}$, and $\textrm{ViT}_{\textrm{WSI}}$ configuration yields the best overall results, highlighting the importance of heterogeneous multi-scale encoders.
}
\label{table2}
\end{table*}

\begin{table*}[t]
\centering
\resizebox{\linewidth}{!}{%
\begin{tabular}{l|c c c|c cccc}
\hline
\multirow{2}{*}{Variants} &  \multicolumn{3}{c|}{$\textrm{ViT}_{\textrm{cell-cell}}$} &PANDA& EBRAINS &WSI-VQA& SlideBench-VQA \\
 & \# Encoder (n) & \# heads (h) &  Dimension (d) & (BA) & (BA)&(A)&(BCNB) (A)\\
\hline
a. MLLM-HWSI   & 2 & 2 & 768 &\textbf{0.748}&\textbf{0.612}&\textbf{0.692}&\textbf{0.687}\\
b. MLLM-HWSI   &4 & 4 & 768 &0.726 &0.592&0.681&\underline{0.677} \\
c. MLLM-HWSI   &6 & 6 & 768 &0.727  &0.590&0.682&0.675 \\
d. MLLM-HWSI   &2 & 2 & 384&\underline{0.741} &0.595&0.688&0.671 \\
e. MLLM-HWSI   &2 & 2 & 192 &0.723&\underline{0.596}&\underline{0.690}&0.676 \\
\hline
\multirow{2}{*}{Variants} &  \multicolumn{3}{c|}{Pooling Operation} &PANDA& EBRAINS & WSI-VQA&SlideBench-VQA \\
 & Max  & Min  & Average & (BA) & (BA) &(A)&(BCNB) (A)\\
\hline
 f. MLLM-HWSI& $\checkmark$& & &0.615&0.521&0.653&0.621\\
 g. MLLM-HWSI&& $\checkmark$ & &0.605&0.545&0.636&0.618\\
 h. MLLM-HWSI&&&$\checkmark$ &0.593&0.543&0.648&0.635\\
 \hline
\end{tabular}
}
\caption{
\textbf{Effect of $\textrm{ViT}_{\textrm{cell-cell}}$ architecture on performance.}
Variants (a–e) modify the number of encoders ($n$), heads ($h$), and embedding dimensions ($d$), while (f–h) use max, min, and average pooling instead of attention. 
Results (BA for PANDA/EBRAINS, A for WSI-VQA/SlideBench-VQA) show that the $n=2$, $h=2$, $d=768$ configuration performs best, emphasizing the value of attention-based cell-level modeling.}
\label{table3}
\end{table*}

\begin{table*}[t!]
\centering
\resizebox{\linewidth}{!}{%
\begin{tabular}{l|llcc|cccc}
\hline
\multirow{2}{*}{Models}&Cell&Patch&Region&WSI&PANDA& EBRAINS &WSI-VQA& SlideBench-VQA\\
&Feat.&Feat.&Feat.&Feat.&\cite{Bulten2022_PANDA} (BA)&\cite{roetzer2022digital} (BA)&\cite{chen2025wsi} (A)&(BCNB) \cite{chen2025slidechat} (A)\\
\hline
WSI-LLaVA \cite{liang2025wsillavamultimodallargelanguage}&$\times$&$\times$&$\times$&$\checkmark$&0.644&0.501&0.546&0.553\\
SlideChat \cite{chen2025slidechat}&$\times$&$\times$&$\times$&$\checkmark$&0.633&0.466&0.601&0.541\\
\hline
MLLM-HWSI$_{1}$&$\times$&$\times$&$\times$&$\checkmark$&0.661&0.519&0.616&0.576\\
MLLM-HWSI$_{2}$&$\times$&$\times$&$\checkmark$&$\checkmark$&0.686&0.534&0.611&0.592\\
MLLM-HWSI$_{3}$&$\times$&$\checkmark$&$\checkmark$&$\checkmark$&0.711&0.566&0.661&0.621\\
MLLM-HWSI$_{4}$&$\checkmark$&$\times$&$\times$&$\checkmark$&0.674&0.531&0.613&0.588\\
MLLM-HWSI$_{5}$&$\times$&$\checkmark$&$\times$&$\checkmark$&0.698&0.548&0.623&0.606\\
MLLM-HWSI$_{6}$&$\checkmark$&$\checkmark$&$\times$&$\checkmark$&\underline{0.715}&0.575&\underline{0.669}&0.640\\
MLLM-HWSI$_{7}$&$\checkmark$&$\times$&$\checkmark$&$\checkmark$&0.714&0.587&0.668&\underline{0.653}\\
MLLM-HWSI&$\checkmark$&$\checkmark$&$\checkmark$&$\checkmark$&\textbf{0.748}&\textbf{0.612}&\textbf{0.692}&\textbf{0.687}\\
\hline
MLLM-HWSI$_{8}$&$\checkmark$&$\times$&$\times$&$\times$&0.616&0.476&0.569&0.522\\
MLLM-HWSI$_{9}$&$\times$&$\checkmark$&$\times$&$\times$&0.623&0.491&0.578&0.521\\
MLLM-HWSI$_{10}$&$\times$&$\times$&$\checkmark$&$\times$&0.631&0.511&0.581&0.529\\
MLLM-HWSI$_{11}$&$\checkmark$&$\checkmark$&$\times$&$\times$&0.675&0.543&0.621&0.577\\
MLLM-HWSI$_{12}$&$\checkmark$&$\times$&$\checkmark$&$\times$&0.672&0.538&0.618&0.574\\
MLLM-HWSI$_{13}$&$\times$&$\checkmark$&$\checkmark$&$\times$&0.673&0.535&0.612&0.566\\
MLLM-HWSI$_{14}$&$\checkmark$&$\checkmark$&$\checkmark$&$\times$&0.712&\underline{0.588}&0.666&0.623\\
\hline
\end{tabular}
}
\caption{
\textbf{Effect of hierarchical representations in MLLM-HWSI.}
Progressive inclusion of cell-, patch-, region-, and WSI-level features in MLLM-HWSI$_{1-3}$ improves performance across all benchmarks.
The full MLLM-HWSI achieves the highest scores, confirming the importance of hierarchical multi-scale alignment.
PANDA and EBRAINS datasets are used for zero-shot classification while WSI-VQA and SlideBench-VQA (BCNB) datasets are used for VQA task.
Feat. stands for ``\textit{Features}'', BA stands for ``\textit{Balanced Accuracy}", and A stands for ``\textit{Accuracy}''.}
\label{table_representation}
\end{table*}

Therefore, the hierarchical WSI–caption alignment mechanism in MLLM-HWSI is central to connecting the visual semantics of histopathology with the descriptive reasoning expressed in diagnostic language \cite{steele2020vocabulary, cheville1983cell, culling2014cellular}. 
In conventional CPath MLLMs \cite{chen2025slidechat, liang2025wsillavamultimodallargelanguage}, caption alignment is performed only at the global level—linking an entire WSI to its corresponding report or summary.
While effective for coarse labeling or WSI-level classification, this approach overlooks the fine-grained relationships between local morphological features and the textual phrases that describe them. 
Hierarchical WSI–caption alignment overcomes this limitation by establishing multi-level correspondences between visual evidence and linguistic descriptions across the full diagnostic hierarchy, enabling precise, interpretable, and clinically coherent visual–language reasoning.

At the representational level, hierarchical caption alignment ensures that visual embeddings from different hierarchical levels—cellular, patch-level, regional, and global—are aligned with language tokens of equivalent semantic granularity. 
Words or short phrases describing morphology (e.g., “hyperchromatic nuclei,” “mitotic figures”) align naturally with cell-level embeddings; sentences describing structural patterns (e.g., “disorganized glandular arrangement”, “stromal invasion”) align with region-level features; and full diagnostic summaries align with the WSI-level representation \cite{chen2022scaling, song2023artificial}. 
This multi-scale correspondence transforms caption generation from a monolithic text synthesis problem into a structured reasoning process, where the model progressively integrates information across scales to compose a coherent narrative of pathology. 
The result is a caption that not only summarizes findings but also reflects how human pathologists articulate diagnostic observations.

From a clinical perspective, hierarchical WSI–caption alignment bridges the gap between machine perception and human explanation \cite{qu2024rise, harada2020molecular}. 
In real-world diagnostic practice, pathologists document their findings hierarchically: starting with cellular morphology, describing architectural context, and concluding with a diagnostic impression \cite{fraggetta2017routine, brixtel2022whole, baidoshvili2023whole, hijazi2024digital}.
For example, a typical breast carcinoma report might read, “\textit{The tumor displays irregular ductal structures lined by pleomorphic epithelial cells with hyperchromatic nuclei and increased mitotic activity}.” 
Each component of this description corresponds to a specific spatial scale within the tissue. 
By aligning these text segments with the respective visual features, MLLM-HWSI enables the model to “\textit{speak the language of pathology}” — generating captions that explicitly refer to verifiable visual evidence.
This interpretability enhances clinical transparency, allowing practitioners to trace each diagnostic statement back to its morphological basis, a critical requirement for medical AI adoption.

On a modeling level, hierarchical caption alignment serves as an additional supervisory signal that strengthens the multi-scale visual–language embedding space.
Aligning visual tokens with hierarchical captions encourages the network to encode features that are both discriminative for diagnosis and descriptive for reporting. 
This dual objective reduces overfitting to classification labels and promotes a richer representation capable of supporting diverse downstream tasks, including report generation, retrieval, and VQA. 
Furthermore, the caption alignment process improves semantic calibration between local and global features: by ensuring that lower-level embeddings contribute meaningfully to higher-level textual synthesis, the model maintains consistency between fine-grained details and WSI-level conclusions.

Empirically, hierarchical WSI–caption alignment enables MLLM-HWSI to produce captions that resemble expert-pathology reports—concise yet semantically dense, containing morphological detail, architectural context, and diagnostic interpretation in a single, coherent paragraph. 
Such outputs demonstrate not only the model’s ability to describe what is visible but also to explain why those features are diagnostically relevant. 
This capability moves beyond simple visual description toward clinically useful, interpretable reasoning, establishing MLLM-HWSI as a bridge between computational pathology and real-world diagnostic reporting.

\section{Additional Ablation Studies}
\label{ablation}
\subsection{Cell Segmentation Backbones (Table~\ref{table1})}
Table~\ref{table1} reports the performance when the backbone cell segmentation method is varied within $\textrm{ViT}_{\textrm{cell-cell}}$. 
The SAM-based CellViT~\cite{horst2024cellvit} achieves the best results.

\subsection{Impact of Different Visual Encoders (Table~\ref{table2})}
Table~\ref{table2} replaces patch/region encoders with UNI, CONCH, or GigaPath, and the WSI encoder with UNI, CONCH, or LongNet, using aggregation layers trained under our losses. 
Homogeneous stacks (all-UNI or all-CONCH) reduce feature diversity and underperform the proposed encoder mix. 
Combining LongNet with CONCH, GigaPath, or UNI improves over homogeneous variants but still lags our proposed configuration.

\subsection{Variants of $\textrm{ViT}_{\textrm{cell-cell}}$ (Table~\ref{table3})}
Table~\ref{table3} studies architectural choices for $\textrm{ViT}_{\textrm{cell-cell}}$: number of encoder blocks $n\!\in\!\{2,4,6\}$, heads $h\!\in\!\{2,4,6\}$, and embedding dimension $d\!\in\!\{768,384,192\}$. 
The configuration $n{=}2$, $h{=}2$, $d{=}768$ yields the best overall results. 
Replacing $\textrm{ViT}_{\textrm{cell-cell}}$ with simple min/max/average pooling leads to significant degradation, indicating the necessity of attention-based cell–cell interaction.

\subsection{Importance of Hierarchical Representations (Table \ref{table_representation})}
As shown in Table~\ref{table_representation}, we progressively augment the hierarchical features in MLLM-HWSI$_{1-3}$. 
Using only WSI-level features (MLLM-HWSI$_1$) already exceeds baseline methods. 
Adding region, patch, and cell-level features yields consistent improvements across all datasets. 
A complementary \emph{subtractive} study (MLLM-HWSI$_{4-7}$) causes notable drops, underscoring the importance of every representation level.

Table~\ref{table_representation} analyzes the contribution of hierarchical representations at different hierarchical levels within MLLM-HWSI.
The variants MLLM-HWSI$_{1-3}$ incrementally incorporate additional levels of hierarchy—starting from WSI-level features alone, then progressively adding region-, patch-, and cell-level embeddings. 
Even with only WSI-level features (MLLM-HWSI$_1$), the model already surpasses strong baselines such as SlideChat and WSI-LLaVA, indicating that the hierarchical pre-training strategy captures rich global contextual features. 
As finer-scale information is introduced, performance consistently improves across all datasets.
The proposed MLLM-HWSI model, which combines cell-, patch-, region-, and WSI-level embeddings, achieves the best overall performance, reaching 74.80$\%$ and 61.20$\%$ balanced accuracy on PANDA and EBRAINS, respectively, and 69.20$\%$ and 68.70$\%$ accuracy on WSI-VQA and SlideBench-VQA.

These gains demonstrate that hierarchical representations allow the model to integrate cellular morphology, microarchitectural context, and global tissue organization into a unified reasoning process. 
The complementary subtractive analysis (MLLM-HWSI${4-7}$) further validates this effect—removing any representation hierarchy leads to a measurable drop in performance, particularly when cell- or patch-level features are excluded, reflecting the importance of fine-grained morphological grounding. 
Models retaining only cell-, patch-, or region-level features (MLLM-HWSI${8-10}$) perform significantly worse, underscoring the necessity of multi-scale contextual integration.

Overall, these results confirm that each hierarchical representation contributes meaningfully to diagnostic accuracy. 
The full MLLM-HWSI, which aligns all four levels of representation, yields the most robust and interpretable performance, emulating how pathologists synthesize information across magnifications—from cellular detail to WSI-level context—to reach precise diagnostic conclusions.

\subsection{Effect of the LLM (Table~\ref{table4})}
Table~\ref{table4} evaluates Vicuna-7B-v1.5~\cite{chiang2023vicuna}, Phi-3-Mini-4k-Instruct~\cite{abdin2024phi}, Llama3-8B-Instruct~\cite{grattafiori2024llama}, InternLM2-Chat-7B~\cite{cai2024internlm2}, and Qwen2-2.5 7B-Instruct~\cite{yang2024qwen2} within MLLM-HWSI. 
Qwen2-2.5 7B-Instruct attains the best performance; the other four are competitive, highlighting the generalization of our framework.

\subsection{Semantic Patch Filtering (SPF) (Table~\ref{table5}-\ref{table6})}
\begin{figure}[t]
\centering
\includegraphics[width=\linewidth]{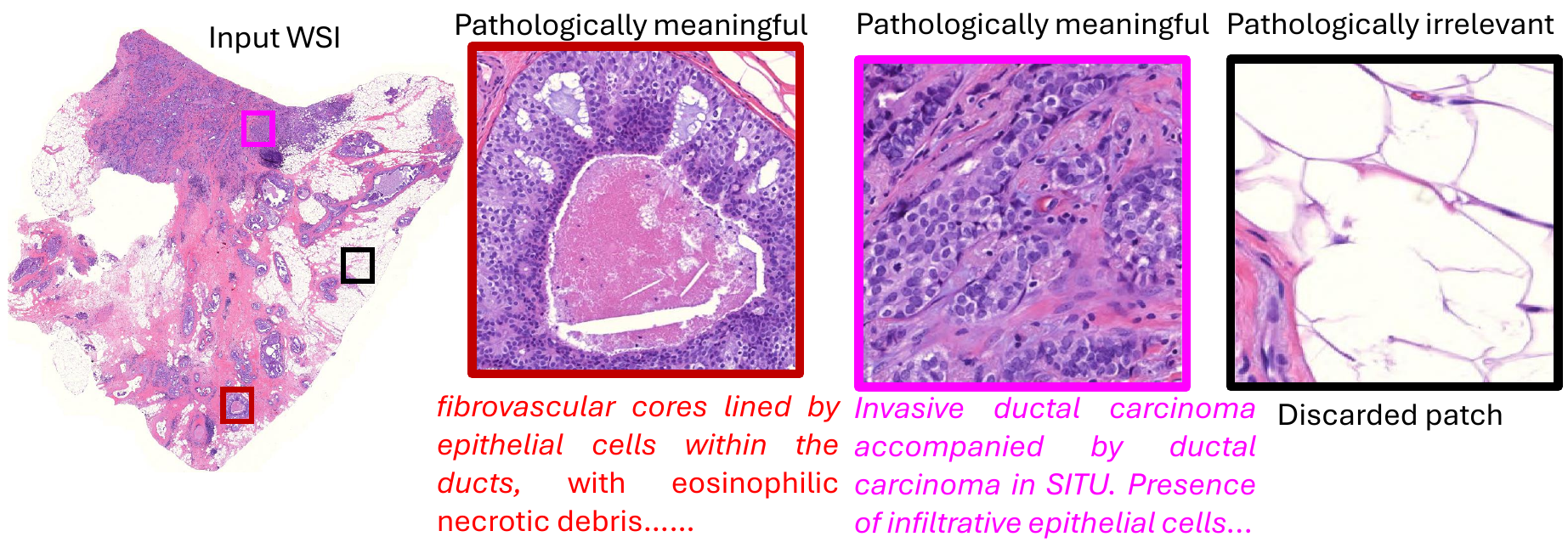}
\caption{Pathologically meaningful patches and discarded patch.}
\label{fig_mean}
\end{figure}

Table~\ref{table5} examines Heterogeneous Patch Selection (HPS) and Diagnostically Relevant Patch Selection (DPS). 
For DPS we select top-$k{=}48$ patches per region $R_i$ (Table \ref{table6}). 
Removing HPS and/or DPS substantially degrades performance; substituting HPS with $k$-means clustering also reduces accuracy. 
Table~\ref{table6} varies the DPS top-$k\!\in\!\{32,64,96\}$ (and additional values), with the best results at $k{=}48$.
Pathologically meaningful qualitative patches are shown in Fig. \ref{fig_mean}.

In our experiments, SPF dynamically selects 48 patches per region (Table~\ref{table5}-\ref{table6}) before LLM input. 
After SPF and cell–cell attention fusion ($\textrm{ViT}_{\textrm{cell-cell}}$), each patch yields one cell and one patch token, each region yields one region token, plus one WSI token, resulting in $\sim$1941 tokens/WSI always below the 2048 token limit, with no truncation.
For a 4096-dim FP16 LLM, this corresponds to $\sim$15 MB of input embeddings and $\sim$30–45 MB total memory, including the KV cache.

\begin{table}[t]
\resizebox{\linewidth}{!}{%
\centering
\begin{tabular}{lccccc}
\hline
\multirow{2}{*}{\textbf{LLMs}} & SlideBench &WSI-&PANDA \\
& (BCNB) (A) &VQA (A)&(BA)\\
\hline
Vicuna-7B-v1.5 \cite{chiang2023vicuna}&\underline{0.682}&0.683&0.725\\
Phi-3-Mini-4k-Instruct \cite{abdin2024phi}&0.664&0.674&0.706\\
Llama3-8B-Instruct \cite{grattafiori2024llama}&0.677&0.673&0.725 \\
Internlm2-Chat-7B \cite{cai2024internlm2}&0.681&\underline{0.684}&\underline{0.738} \\
Qwen2-2.5 7B-Instruct \cite{yang2024qwen2}&\textbf{0.687}&\textbf{0.692}&\textbf{0.748}\\
\hline
\end{tabular}
}
\caption{\textbf{Effect of LLM choice on VQA performance.}
Comparison of five instruction-tuned LLMs integrated into MLLM-HWSI across SlideBench (BCNB), WSI-VQA, and PANDA datasets. 
Qwen2.5-7B-Instruct yields the highest accuracy, highlighting its stronger multimodal reasoning capability.}
\label{table4}
\end{table}

\begin{table*}[t]
\resizebox{\linewidth}{!}{%
\centering
\begin{tabular}{l|c c |c cccc}
\hline
\multirow{2}{*}{Variants} &  \multicolumn{2}{c|}{Semantic Patch Filtering} &PANDA &EBRAINS& WSI-VQA& SlideBench-VQA \\
 &~~~~HPS&~~~~DPS &(BA)&(BA)&(A)&(BCNB)(A)\\
\hline
a. MLLM-HWSI   & $\checkmark$ &  $\checkmark$ &\textbf{0.748}&\textbf{0.612}&\textbf{0.692}&\textbf{0.687}\\
b. MLLM-HWSI   & $\checkmark$ &  $\times$ &0.731&0.592&0.674&0.663\\
c. MLLM-HWSI   & $\times$ &  $\checkmark$ &\underline{0.741}&\underline{0.606}&\underline{0.684}&\underline{0.676}\\
d. MLLM-HWSI   & $\times$ &  $\times$ &0.711&0.566&0.664&0.657\\
\hline
\multirow{2}{*}{Variants} &  \multicolumn{2}{c|}{Semantic Patch Filtering} &PANDA& EBRAINS& WSI-VQA&SlideBench-VQA \\
 &~~~~K-means&~~~~DPS & (BA) & (BA) &(A) &(BCNB) (A)\\
e. MLLM-HWSI   & $\checkmark$ &  $\checkmark$ &0.702&0.571&0.661&0.654\\
f. MLLM-HWSI   & $\checkmark$ &  $\times$ &0.683&0.554&0.641&0.644\\
 \hline
\end{tabular}
}
\caption{\textbf{Effect of Semantic Patch Filtering.}
Comparison of different combinations of Heterogeneous Patch Selection (HPS), Diagnostically Relevant Patch Selection (DPS), and K-means clustering in MLLM-HWSI. 
The combination of HPS and DPS yields the best overall accuracy, highlighting their complementary roles in selecting diverse and diagnostic patches.}
\label{table5}
\end{table*}

\begin{table}[h!]
\resizebox{\columnwidth}{!}{%
\centering
\begin{tabular}{l|llcc|}
\hline
top-$k$&PANDA& EBRAINS &WSI-VQA& SlideBench-VQA\\
Value&\cite{Bulten2022_PANDA} (BA)&\cite{roetzer2022digital} (BA)&\cite{chen2025wsi} (A)&(BCNB) \cite{chen2025slidechat}(A)\\
\hline
16&0.711&0.588&0.676&0.641\\
32&0.731&0.605&0.681&0.663\\
48&\textbf{0.748}&\textbf{0.612}&\textbf{0.692}&\textbf{0.687}\\
64 &\underline{0.743}&0.604&\underline{0.686}&\underline{0.682}\\
96&0.740&0.608&0.685&0.678\\
128&0.735&\underline{0.610}&0.684&0.675\\
\hline
\end{tabular}
}
\caption{
\textbf{Influence of top-$k$ in $\textrm{ViT}_{\textrm{cell-cell}}$.}
Performance with different top-$k$ values in the Diagnostically Relevant Patch Selection (DPS) module.  
The best results are achieved at top-$k=48$, indicating optimal diagnostic coverage and compactness.
}
\label{table6}
\end{table}

\section{Computational Complexity}
\label{time}

The model was implemented on four NVIDIA A100 GPUs.  
During zero-shot inference, MLLM-HWSI required an average of 4.90 minutes per WSI on the BRAINS30 dataset, compared to 4.3, 4.4, and 3.8 minutes for SlideChat, TITAN, and WSI-LLaVA, respectively.  
The additional time arises from multi-scale feature extraction and semantic patch filtering, which enhance performance at a modest computational cost.  
\textit{Despite incorporating hierarchical multi-scale feature extraction, MLLM-HWSI maintains computational efficiency comparable to existing SOTA models, demonstrating scalability without significant inference overhead.}

\section{WSI-level Classification Results}
\label{wsiresults}
\subsection{Zero-shot Classification of WSIs (Table \ref{tab:wsi_zeroshot})}
We evaluated the zero-shot WSI classification capability of the pre-trained MLLM-HWSI model using the vision and text encoders obtained from Stage I (hierarchical cross-modal alignment). Following established evaluation protocols in TITAN \cite{ding2024titan}, CONCH \cite{lu2024visual}, and QuiltNet \cite{ikezogwo2024quilt}, we directly measured the semantic alignment between hierarchical WSI features and class-specific textual descriptions without any task-specific fine-tuning.

For each test WSI, hierarchical visual features were extracted from the MLLM-HWSI encoder and compared against class-level textual prompts encoded by the text encoder. 
Both visual and textual embeddings were $\ell_{2}$-normalized, and class prediction was determined by selecting the label corresponding to the highest cosine similarity between the two modalities. 
We adopted dataset-specific testing prompts consistent with prior zero-shot WSI classification works to ensure fair comparison across benchmarks \cite{ding2024titan, lu2024visual, ikezogwo2024quilt, chen2024towards}.

This protocol evaluates how effectively MLLM-HWSI transfers its learned hierarchical alignment from multimodal pre-training to unseen classification tasks.
As shown in Fig. 4(a) of the main paper and Table \ref{tab:wsi_zeroshot}, MLLM-HWSI achieves SOTA zero-shot accuracy across six external datasets, demonstrating robust generalization and the discriminative strength of its multi-scale visual–language representations.

\begin{table*}[t!]
\centering
\caption{WSI-level Zero-shot classification performance comparison results with SOTA CPath models across six datasets.}
\resizebox{\textwidth}{!}{
\begin{tabular}{l|cc|cc|cc|cc|cc|cc}
\hline
\textbf{Method} & \multicolumn{2}{c|}{\textbf{PANDA}} & \multicolumn{2}{c|}{\textbf{EBRAINS}} & \multicolumn{2}{c|}{\textbf{BRACS}} & \multicolumn{2}{c|}{\textbf{UBC-Ocean}} & \multicolumn{2}{c|}{\textbf{TCGA-OT}} & \multicolumn{2}{c}{\textbf{IMP-CRC}} \\
& F & BA & F & BA & F & BA & F & BA & F & BA & F & BA \\
\hline
PLIP & 0.288 & 0.235 & 0.013 & 0.080 & 0.214 & 0.203 & 0.376 & 0.345 & 0.203 & 0.223 & 0.523 & 0.655 \\
PathCLIP & 0.461 & 0.455 & 0.223 & 0.187 & 0.281 & 0.309 & 0.657 & 0.612 & 0.304 & 0.334 & 0.560 & 0.700 \\
CPLIP & 0.445 & 0.420 & 0.253 & 0.233 & 0.294 & 0.288 & 0.706 & 0.653 & 0.431 & 0.405 & 0.591 & 0.733 \\
CONCH & 0.596 & 0.566 & 0.304 & 0.278 & 0.336 & 0.344 & 0.786 & 0.807 & 0.488 & 0.532 & 0.637 & 0.833 \\
QuiltNet & 0.532 & 0.509 & 0.229 & 0.201 & 0.321 & 0.312 & 0.753 & 0.776 & 0.486 & 0.506 & 0.608 & 0.788 \\
MR-PLIP & \underline{0.701} & \textbf{0.681} & 0.332 & 0.314 & \underline{0.403} & \underline{0.411} & 0.855 & 0.833 & 0.506 & 0.541 & \underline{0.679} & 0.809 \\
SlideChat & 0.605 & 0.633 & 0.479 & 0.326 & 0.248 & 0.255 & 0.861 & \underline{0.902} & 0.493 & 0.487 & 0.648 & 0.809 \\
PathGenCLIP & 0.511 & 0.488 & 0.255 & 0.221 & 0.295 & 0.288 & 0.786 & 0.756 & 0.498 & 0.522 & 0.612 & 0.723 \\
MI-Zero & 0.405 & 0.386 & 0.253 & 0.233 & 0.261 & 0.241 & 0.807 & 0.786 & 0.506 & 0.486 & 0.585 & 0.666 \\
KEP & 0.476 & 0.455 & 0.209 & 0.193 & 0.244 & 0.221 & 0.734 & 0.721 & 0.446 & 0.456 & 0.598 & 0.687 \\
TITAN & 0.621 & 0.608 & \underline{0.365} & \underline{0.543} & 0.385 & 0.381 & \underline{0.908} & 0.865 & \underline{0.713} & \underline{0.616} & \underline{0.723} & \underline{0.861} \\
WSI-LLaVA & 0.668 & 0.644 & 0.389 & 0.501 & 0.294 & 0.289 & 0.872 & 0.881 & 0.517 & 0.523 & 0.692 & 0.823 \\
PRISM & 0.544 & 0.521 & 0.263 & 0.279 & 0.322 & 0.334 & 0.753 & 0.765 & 0.538 & 0.460 & 0.624 & 0.743 \\
\hline
\textbf{Proposed MLLM-HWSI} & \textbf{0.722} & \textbf{0.748} & \textbf{0.458} & \textbf{0.612} & \textbf{0.446} & \textbf{0.456} & \textbf{0.952} & \textbf{0.922} & \textbf{0.748} & \textbf{0.666} & \textbf{0.767} & \textbf{0.908} \\
\hline
\end{tabular}
}
\label{tab:wsi_zeroshot}
\end{table*}

\subsection{Linear Probe Evaluation (Table \ref{tab:wsi_linear_probe})}
We also conducted a linear probe evaluation to assess the discriminative strength and transferability of the representations learned by MLLM-HWSI during pre-training. 
Linear probing provides a widely adopted, architecture-agnostic framework for measuring the quality of learned features \cite{ding2024titan, chen2024towards}. 
The procedure involves freezing all parameters of the pre-trained encoder and training a simple logistic regression classifier on the extracted features. 
High linear probe performance indicates that the encoder captures rich, separable, and generalizable representations.
Please see our linear probe evaluation results in Fig. 4 (b) of the main manuscript and Table \ref{tab:wsi_linear_probe}.

Following prior CPath foundation models such as TITAN \cite{ding2024titan} and UNI \cite{chen2024towards}, we trained a linear classifier on top of hierarchical features extracted from the Stage I MLLM-HWSI encoder. 
The classifier was optimized using an $\ell_2$-regularized L-BFGS solver from \texttt{scikit-learn}, with a maximum of 500 iterations. 
For datasets lacking a dedicated validation set, we used default settings with $\ell_2 = 1$ and 1,000 iterations to ensure stable convergence. 
The linear classifier was trained using cross-entropy loss on frozen embeddings aggregated across cell-, patch-, region-, and slide-level tokens.

Table \ref{tab:wsi_linear_probe} presents results across six public datasets, comparing MLLM-HWSI to leading CPath foundation models, including TITAN, FOCUS, GigaPath, and UNI. O
MLLM-HWSI model consistently achieves the best performance across all datasets and metrics, attaining the highest F1-score (F) and balanced accuracy (BA) on PANDA (0.882 / 0.867), EBRAINS30 (0.833 / 0.803), BRACS (0.603 / 0.571), UBC-Ocean (0.968 / 0.961), TCGA-OT (0.789 / 0.766), and IMP-CRC (0.951 / 0.981). 
These substantial improvements over strong baselines such as TITAN (0.836 / 0.823 on PANDA) and UNI (0.809 / 0.757 on PANDA) demonstrate that hierarchical vision–language alignment yields highly discriminative and transferable WSI representations.
Overall, the linear probe results confirm that MLLM-HWSI learns semantically structured, multi-scale embeddings that generalize effectively across organs, cancer types, and dataset domains—validating the effectiveness of hierarchical pre-training in capturing biologically meaningful and diagnostic features.

\begin{table*}[t!]
\centering
\caption{WSI-level classification results and comparisons using linear probe evaluation and weakly supervised MIL-based classification with SOTA CPath models across six datasets.}
\resizebox{\textwidth}{!}{
\begin{tabular}{l|cc|cc|cc|cc|cc|cc}
\hline
\textbf{Method} &
\multicolumn{2}{c|}{\textbf{PANDA}} &
\multicolumn{2}{c|}{\textbf{EBRAINS30}} &
\multicolumn{2}{c|}{\textbf{BRACS}} &
\multicolumn{2}{c|}{\textbf{UBC-Ocean}} &
\multicolumn{2}{c|}{\textbf{TCGA-OT}} &
\multicolumn{2}{c}{\textbf{IMP-CRC}} \\
\quad & F & BA & F & BA & F & BA & F & BA & F & BA & F & BA \\
\hline
HIPT & 0.687 & 0.654 & 0.702 & 0.677 & 0.334 & 0.288 & 0.766 & 0.706 & 0.512 & 0.488 & 0.718 & 0.801 \\
CTransPath & 0.752 & 0.691 & 0.597 & 0.514 & 0.398 & 0.355 & 0.788 & 0.733 & 0.566 & 0.544 & 0.749 & 0.833 \\
REMEDIS & 0.766 & 0.711 & 0.471 & 0.382 & 0.367 & 0.331 & 0.733 & 0.706 & 0.504 & 0.455 & 0.772 & 0.843 \\
CHIEF & 0.745 & 0.724 & 0.706 & 0.688 & 0.413 & 0.387 & 0.823 & 0.789 & 0.640 & 0.528 & 0.701 & 0.781 \\
DinoPath & 0.682 & 0.706 & 0.771 & \underline{0.755} & 0.394 & 0.361 & 0.844 & 0.821 & 0.586 & 0.556 & 0.792 & 0.855 \\
Virchow & 0.741 & 0.728 & 0.723 & 0.701 & 0.459 & 0.422 & 0.902 & 0.889 & 0.656 & 0.564 & 0.708 & 0.755 \\
RudolfV & 0.653 & 0.677 & 0.706 & 0.688 & 0.438 & 0.401 & 0.881 & 0.865 & 0.607 & 0.596 & 0.732 & 0.786 \\
UNI & 0.809 & 0.757 & 0.746 & 0.675 & \underline{0.538} & \underline{0.504} & 0.940 & 0.922 & 0.657 & 0.633 & 0.814 & 0.881 \\
GigaPath & 0.789 & 0.794 & 0.704 & 0.687 & 0.507 & 0.477 & 0.901 & 0.889 & 0.659 & 0.543 & 0.791 & 0.856 \\
TITAN & \underline{0.836} & \underline{0.823} & \underline{0.786} & 0.735 & 0.511 & 0.400 & \underline{0.956} & \underline{0.933} & \underline{0.764} & \underline{0.704} & \underline{0.903} & \underline{0.946} \\
FOCUS & 0.804 & 0.782 & 0.733 & 0.671 & 0.474 & 0.451 & 0.903 & 0.841 & 0.685 & 0.605 & 0.829 & 0.877 \\
\hline
\textbf{Proposed MLLM-HWSI} & \textbf{0.882} & \textbf{0.867} & \textbf{0.833} & \textbf{0.803} & \textbf{0.603} & \textbf{0.571} & \textbf{0.968} & \textbf{0.961} & \textbf{0.789} & \textbf{0.766} & \textbf{0.951} & \textbf{0.981} \\
\hline
\end{tabular}
}
\label{tab:wsi_linear_probe}
\end{table*}

\section{WSI-Level Report Generation Qualitative Results (Tables  \ref{tab:report_comparison1}-\ref{tab:report_comparison5})}
\label{qualitative}
We conducted an extensive qualitative comparison of pathology report generation to evaluate the interpretive and diagnostic reasoning capabilities of MLLM-HWSI against SOTA CPath models, including WSI-LLaVA, MI-Gen, Hist-Gen, Quilt-LLaVA, and GPT-4o. 
Tables \ref{tab:report_comparison1}–\ref{tab:report_comparison5} illustrate representative examples covering multiple diagnostic contexts—morphological description, global architecture analysis, key diagnostic feature identification, molecular subtyping, and TNM staging.

Across all examples, MLLM-HWSI produces reports that are nearly indistinguishable from expert-authored ground truth, demonstrating close semantic and morphological alignment.
Its outputs consistently capture fine-grained histological detail—including nuclear pleomorphism, keratinization, intercellular bridges, and mitotic figures—while preserving global structural context, such as tumor organization and invasion patterns.
The generated descriptions are linguistically coherent, clinically interpretable, and free from redundant or hallucinated content that often appears in baseline models.

In morphological and global description tasks (Tables \ref{tab:report_comparison1}–\ref{tab:report_comparison2}), MLLM-HWSI accurately describes both cellular morphology and tissue-level architecture, surpassing prior models that either miss key features or overgeneralize findings. 
For diagnostic and molecular interpretation (Tables \ref{tab:report_comparison3}–\ref{tab:report_comparison4}), the model correctly identifies defining histologic and molecular attributes, such as papillary architecture, psammoma bodies, and HPV-negative subtypes, aligning precisely with ground-truth annotations. 
In the staging example (Table \ref{tab:report_comparison5}), MLLM-HWSI achieves perfect correspondence with clinical staging guidelines, correctly reporting T3 N2 M0 without deviation.

Overall, these qualitative analyses highlight that MLLM-HWSI not only surpasses all competing models in accuracy and language fluency but also demonstrates clinically grounded, evidence-based reasoning.
By aligning hierarchical WSI features with pathology-specific language, MLLM-HWSI generates diagnostic narratives that faithfully replicate expert interpretation—bridging the gap between automated analysis and human-level pathological reporting.

\begin{table*}[t!]
\centering
\caption{Qualitative comparison of pathology report generation across SOTA CPath models. 
The qualitative analysis illustrates how MLLM-HWSI produces reports that closely match expert-annotated ground truth, capturing both fine-grained cellular morphology and global architectural context. Compared with prior models (e.g., WSI-LLaVA, MI-Gen, Hist-Gen, Quilt-LLaVA, and GPT-4o), MLLM-HWSI generates linguistically coherent and diagnostically accurate descriptions that mirror expert reasoning, demonstrating superior grounding between visual evidence and clinical language. Green: matched ground-truth content; Red: deviations; Orange: ground truth content missing in model response; Underlined: template language.}
\label{tab:report_comparison1}
\renewcommand{\arraystretch}{1.3}
\resizebox{0.8\textwidth}{!}{%
\begin{tabularx}{\textwidth}{lX}
\toprule
\textbf{WSI} & 
\includegraphics[width=0.25\textwidth]{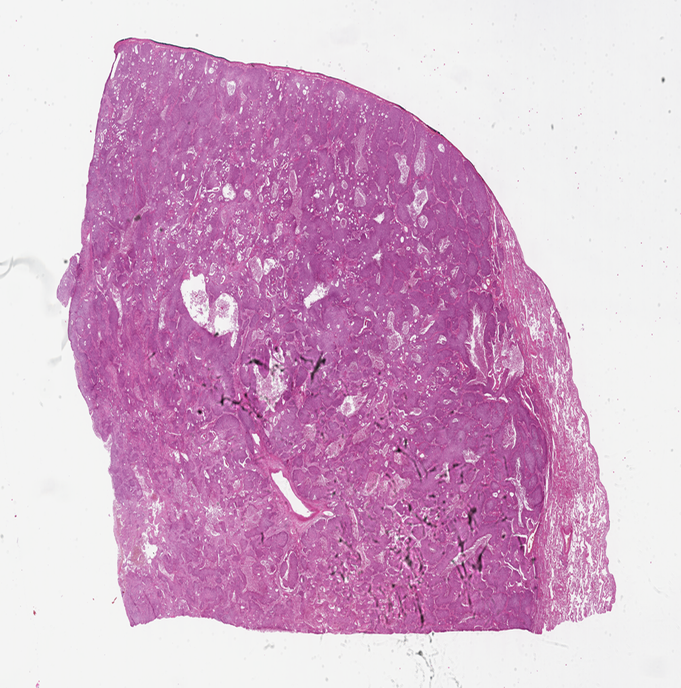} \\

\midrule
\textbf{Question} &
What are the main observations in the whole slide histopathology image, and what is the likely diagnosis? \\

\midrule
\textbf{WSI-LLaVA} &
Microscopic observation of the pathology slide reveals \textcolor{DarkGreen}{moderately differentiated squamous cell carcinoma characterized by
nests and sheets of atypical squamous epithelial cells with intercellular bridges and keratinization. The tumor cells exhibit
moderate nuclear pleomorphism with enlarged, hyperchromatic nuclei and prominent nucleoli. Mitotic figures are present,} \textcolor{red}{
indicating active cell division.}\textcolor{DarkGreen}{ There is no evidence of vascular invasion, necrosis, or calcification within the examined
tissue. Based on these observations, the diagnosis is moderately differentiated keratinizing squamous carcinoma.} \\

\midrule
\textbf{MI-Gen} &
Microscopic observation of the pathology slide reveals a \textcolor{red}{poorly} \textcolor{DarkGreen}{ differentiated squamous cell carcinoma characterized} by a
lack of squamous cell carcinoma. The tumor exhibits a lack of squamous differentiation with a lack of squamous
differentiation. \textcolor{DarkGreen}{The tumor cells exhibit moderate nuclear pleomorphism with variability in nuclear size and shape and there
is a moderate rate of cell division. Vascular invasion is not identified. Necrosis is not observed within the tumor. Based on
these observations, the diagnosis is squamous cell carcinoma. }\\

\midrule
\textbf{Hist-Gen} &
Microscopic observation of the pathology slide reveals a \textcolor{red}{poorly} \textcolor{DarkGreen}{differentiated squamous cell carcinoma characterized by
irregular nests and sheets of atypical squamous cells with intercellular bridges and keratinization.} \textcolor{red}{The tumor cells exhibit a
high degree of nuclear pleomorphism.} \textcolor{DarkGreen}{There is no evidence of vascular invasion or necrosis within the tumor tissue.} The
surrounding lung parenchyma shows areas of chronic inflammation and chronic inflammation. Based on these observations,
the diagnosis is \textcolor{DarkGreen}{ squamous cell carcinoma}, \textcolor{red}{poorly differentiated.} \\

\midrule
\textbf{Quilt-LLaVA} &
Microscopic observation of the pathology slide reveals \textcolor{DarkGreen}{ sheets and nests of atypical squamous cells with moderate
differentiation. There is evidence of focal keratinization, characterized by the presence of keratin pearls within the tumor
tissue. Intercellular bridges are noted among the squamous cells, indicating differentiation towards squamous epithelium.
The cellular morphology shows moderate pleomorphism, with variability in nuclear size and shape, and occasional
prominent nucleoli. Mitotic figures are present but not excessively numerous. Vascular invasion is not specified in the
observations, and necrosis is not prominently noted. Based on these observations, the diagnosis is squamous cell carcinoma,
moderately differentiated.} \\

\midrule
\textbf{GPT-4o} &
\textcolor{red}{This histopathology image shows a well-circumscribed, nodular lesion with a dense, cellular stroma. There are areas of
necrosis and hemorrhage, and the tissue architecture appears disrupted. The lesion has a heterogeneous appearance with
varying cellular densities and some fibrotic regions. \textbf{Diagnosis:} The features are consistent with a uterine fibroid
(leiomyoma), which is characterized by a smooth muscle cell growth forming a circumscribed mass with areas of
degeneration. }\\

\midrule
\textbf{MLLM-HWSI} & 
\textcolor{DarkGreen}{\underline{Microscopic observation of the pathology slide reveals sheets and nests of atypical squamous epithelial} \underline{cells with moderate differentiation.} There is clear evidence of focal keratinization, highlighted by the formation of keratin pearls within the tumor tissue. Intercellular bridges are well identified among the tumor cells, confirming differentiation toward squamous epithelium. The nuclei display moderate pleomorphism with variation in size and shape and occasional prominent nucleoli. Mitotic figures are present but not numerous, indicating a moderate proliferative index. There is no evidence of vascular invasion, and necrosis is not apparent in the examined sections. \underline{Based on these findings, the features are consistent with a moderately differentiated keratinizing} \underline{squamous cell carcinoma.}}
\\

\midrule
\textbf{Ground Truth} &
Microscopic observation of the pathology slide reveals sheets and nests of atypical squamous cells with moderate differentiation. There is evidence of focal keratinization, characterized by the presence of keratin pearls within the tumor tissue. Intercellular bridges are noted among the squamous cells, indicating differentiation towards squamous epithelium.
The cellular morphology shows moderate pleomorphism, with variability in nuclear size and shape, and occasional prominent nucleoli. Mitotic figures are present but not excessively numerous. Vascular invasion is not specified in the observations, and necrosis is not prominently noted. Based on these observations, the diagnosis is squamous cell carcinoma,
moderately differentiated. \\

\bottomrule
\end{tabularx}
}
\end{table*}

\begin{table*}[htbp]
\centering
\caption{A comparative example of global morphology description outputs from different CPath models.
Green: matched ground-truth content; Red: deviations; Orange: ground truth content missing in model response; Underlined: template language.}
\label{tab:report_comparison2}
\renewcommand{\arraystretch}{1.3}
\resizebox{0.8\textwidth}{!}{%
\begin{tabularx}{\textwidth}{lX}
\toprule
\textbf{WSI} & 
\includegraphics[width=0.25\textwidth]{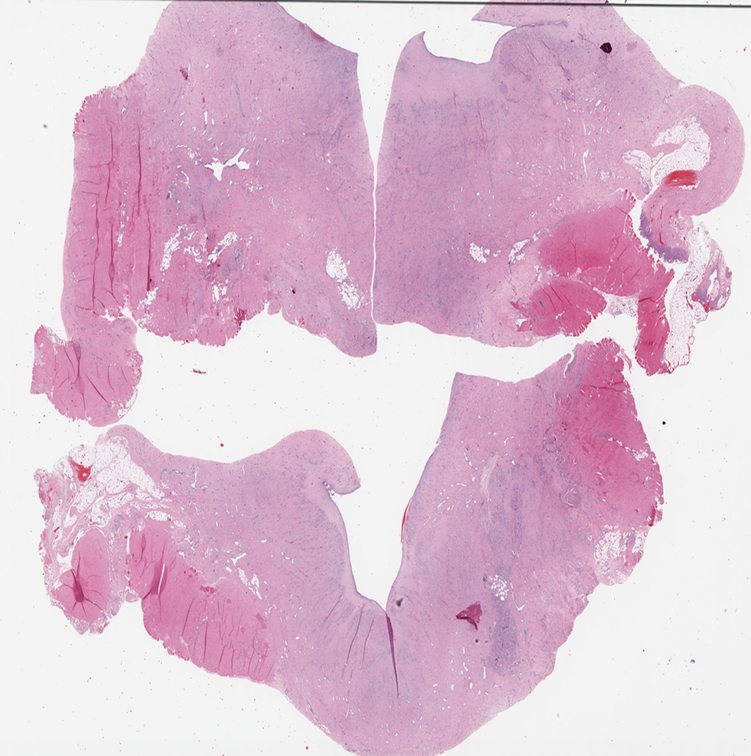} \\

\midrule
\textbf{Question} &
Can you provide an overall description of the whole slide image (WSI)?\\

\midrule
\textbf{WSI-LLaVA} &
\textcolor{DarkGreen}{The whole slide image reveals exuberant granulation tissue characterized by notable fibroblastic proliferation and myxoid changes. The cellular component is primarily composed of spindle-shaped cells within a myxoid matrix. There is an absence of normal tissue architecture as the granulation tissue dominates the observed section. Importantly, no residual sarcoma tissue is identifiable within this examined section.}\\

\midrule
\textbf{Quilt-LLaVA} &
\textcolor{red}{The whole slide image reveals a tumor with a diffuse architectural pattern, characterized by a mixture of alveolar formations and glandular cell differentiation. There is a notable presence of large cells with prominent nucleoli and moderate nuclear atypia. The mitotic activity is moderate, with 8 mitoses per 10 high power fields. The tumor shows significant vascular invasion, with tumor cells infiltrating the renal vein. Additionally, there is focal necrosis within the tumor tissue.} \\

\midrule
\textbf{WSI-VQA} &
\textcolor{red}{The whole slide image reveals a biphasic tumor composed of spindle-shaped cells arranged in a storiform pattern.} The spindle cells are arranged in a storiform pattern. The spindle cells are arranged in a storiform pattern. There is a notable absence of necrosis or calcification within the tumor. 
\\

\midrule
\textbf{GPT-4o} &
The whole slide image appears to be a tissue section stained with hematoxylin and eosin (H\&E), commonly used in histopathology. The sample is divided into several sections, likely representing different orientations or parts of the tissue. -- *Coloration*: The pink and purple hues suggest typical H\&E staining, where eosin stains cytoplasmic components pink and hematoxylin stains nuclear material blue or purple. -- *Tissue Architecture*: The sections show varying thickness with folded tissue smooth and irregular contours, indicating possible different tissue types or regions. -- *Other Observations*: There are areas that seem to contain more densely packed cells and some regions with more open or less cellular areas, suggesting different tissue densities or types. A detailed examination at higher magnification would be necessary for further analysis, such as identifying specific pathological features or tissues. \\

\midrule
\textbf{MLLM-HWSI} &
\textcolor{DarkGreen}{The WSI demonstrates exuberant granulation tissue with extensive fibroblastic proliferation embedded in a myxoid stroma.}
\textcolor{DarkGreen}{The spindle-shaped fibroblasts are loosely arranged within the extracellular matrix, accompanied by delicate capillary networks and scattered inflammatory infiltrates.}
\textcolor{DarkGreen}{No evidence of residual malignant or sarcomatous components is observed.}
\textcolor{DarkGreen}{The overall architecture is dominated by reactive granulation tissue replacing normal parenchymal organization, consistent with post-treatment reparative changes.}

\\

\midrule
\textbf{Ground Truth} &
The whole slide image demonstrates exuberant granulation tissue characterized by notable fibroblastic proliferation and myxoid changes. The cellular component is primarily composed of spindle-shaped cells within a myxoid matrix. There is an absence of normal tissue architecture as the granulation tissue dominates the observed section. Importantly, no residual sarcoma tissue is identifiable within this examined section. \\

\bottomrule
\end{tabularx}
}
\end{table*}

\begin{table*}[htbp]
\centering
\caption{A comparative example of key diagnostic description outputs from different CPath models.
Green: matched ground-truth content; Red: deviations; Orange: ground truth content missing in model response; Underlined: template language.}
\label{tab:report_comparison3}
\renewcommand{\arraystretch}{1.3}
\resizebox{0.8\textwidth}{!}{%
\begin{tabularx}{\textwidth}{lX}
\toprule
\textbf{WSI} & 
\includegraphics[width=0.25\textwidth]{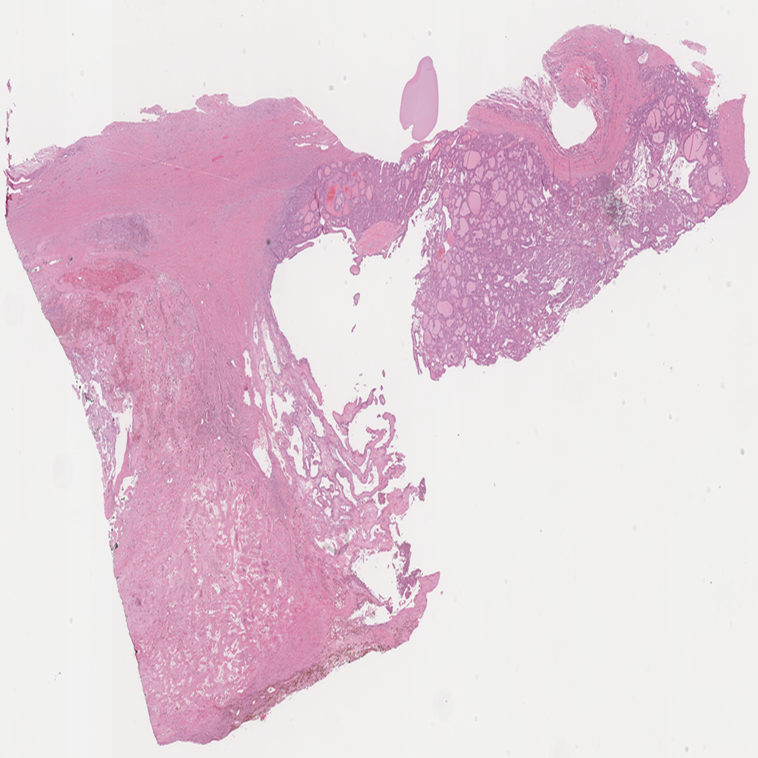} \\

\midrule
\textbf{Question} &
Which observed features are most significant for making a diagnosis?\\

\midrule
\textbf{WSI-LLaVA} &
\textcolor{DarkGreen}{ The most significant features include the papillary architecture with fibrovascular cores, the presence of psammoma bodies, and the nuclear characteristics such as overlapping, grooves, and intranuclear cytoplasmic inclusions. The absence of vascular invasion, necrosis, and calcification also plays a crucial role in the diagnostic process.}\\

\midrule
\textbf{Quilt-LLaVA} &
\textcolor{DarkGreen}{The most significant features include the irregular clusters of epithelial cells} with poor glandular differentiation, \textcolor{DarkGreen}{marked nuclear pleomorphism}, and the presence of perineural invasion. The high mitotic activity and evidence of \textcolor{DarkGreen}{lymphovascular invasion} are also critical, as they indicate aggressive tumor behavior.\\

\midrule
\textbf{WSI-VQA} &
\textcolor{DarkGreen}{The most significant features include the papillary architecture with fibrovascular cores} and the presence of fibrovascular cores. \textcolor{DarkGreen}{The absence of vascular invasion and necrosis further supports the diagnosis. }The absence of necrosis and calcification further supports the diagnosis.
\\

\midrule
\textbf{GPT-4o} &
The whole slide image appears to be a tissue section stained with hematoxylin and eosin (H\&E), commonly used in histopathology. The sample is divided into several sections, likely representing different orientations or parts of the tissue. -- *Coloration*: The pink and purple hues suggest typical H\&E staining, where eosin stains cytoplasmic components pink and hematoxylin stains nuclear material blue or purple. -- *Tissue Architecture*: The sections show varying thickness with folded tissue smooth and irregular contours, indicating possible different tissue types or regions. -- *Other Observations*: There are areas that seem to contain more densely packed cells and some regions with more open or less cellular areas, suggesting different tissue densities or types. A detailed examination at higher magnification would be necessary for further analysis, such as identifying specific pathological features or tissues. \\

\midrule
\textbf{MLLM-HWSI} & 
\textcolor{DarkGreen}{The diagnostic features most critical for interpretation include a prominent papillary architecture with well-formed fibrovascular cores lined by epithelial cells showing characteristic nuclear overlapping, grooves, and intranuclear cytoplasmic inclusions.}
\textcolor{DarkGreen}{Multiple psammoma bodies are present within the papillary stalks.}
\textcolor{DarkGreen}{There is no evidence of vascular, lymphatic, or perineural invasion, and no areas of necrosis are identified.}
\textcolor{DarkGreen}{These cytologic and architectural features are diagnostic of a papillary-patterned neoplasm consistent with papillary carcinoma morphology.}\\
\\
\midrule
\textbf{Ground Truth} &
The whole slide image demonstrates exuberant granulation tissue characterized by notable fibroblastic proliferation and myxoid changes. The cellular component is primarily composed of spindle-shaped cells within a myxoid matrix. There is an absence of normal tissue architecture as the granulation tissue dominates the observed section. Importantly, no residual sarcoma tissue is identifiable within this examined section. \\

\bottomrule
\end{tabularx}
}
\end{table*}

\begin{table*}[htbp]
\centering
\caption{A comparative example of molecular subtyping outputs from different CPath models.}
\label{tab:report_comparison4}
\renewcommand{\arraystretch}{1.3}
\resizebox{0.8\textwidth}{!}{%
\begin{tabularx}{\textwidth}{lX}
\toprule
\textbf{WSI} & 
\includegraphics[width=0.25\textwidth]{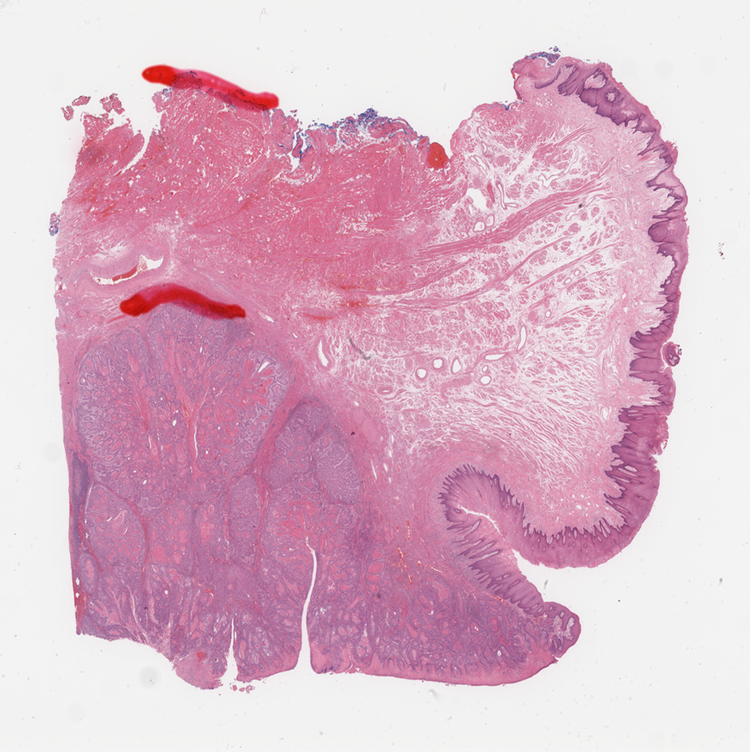} \\

\midrule
\textbf{Question} &
Based on the IHC results showing P16 negativity and HR, HPV ISH status negativity, what is the molecular subtype in this head and neck squamous cell carcinoma?\\

\midrule
\textbf{WSI-LLaVA} &
The molecular subtype is \textcolor{DarkGreen}{HNSC HPV.}\\

\midrule
\textbf{Quilt-LLaVA} &
The molecular subtype is \textcolor{DarkGreen}{HNSC HPV.}\\

\midrule
\textbf{WSI-VQA} &
the molecular subtype is \textcolor{red}{hnsc unk}.
\\

\midrule
\textbf{GPT-4o} &
\textcolor{red}{The molecular subtype of head and neck squamous cell carci
noma with P16 negativity and HR HPV ISH negativity is typ
ically HPV-negative. This subtype is generally associated with non-viral risk factors such as smoking and alcohol use.} \\

\midrule
\textbf{MLLM-HWSI} &
\textcolor{DarkGreen}{Given the P16 negativity and HR HPV ISH negativity, the molecular subtype is correctly classified as HNSC HPV-negative.} 
\textcolor{DarkGreen}{This reflects a non-viral, keratinizing phenotype often linked to tobacco and alcohol-related carcinogenesis.} 
\textcolor{DarkGreen}{These molecular features are characteristic of the HPV-independent pathway in head and neck squamous cell carcinoma.}\\

\\

\midrule
\textbf{Ground Truth} &
The molecular subtype is HNSC HPV\\

\bottomrule
\end{tabularx}
}
\end{table*}

\begin{table*}[htbp]
\centering
\caption{A comparative example of staging outputs from different CPath models.}
\label{tab:report_comparison5}
\renewcommand{\arraystretch}{1.3}
\resizebox{0.8\textwidth}{!}{%
\begin{tabularx}{\textwidth}{lX}
\toprule
\textbf{WSI} & 
\includegraphics[width=0.25\textwidth]{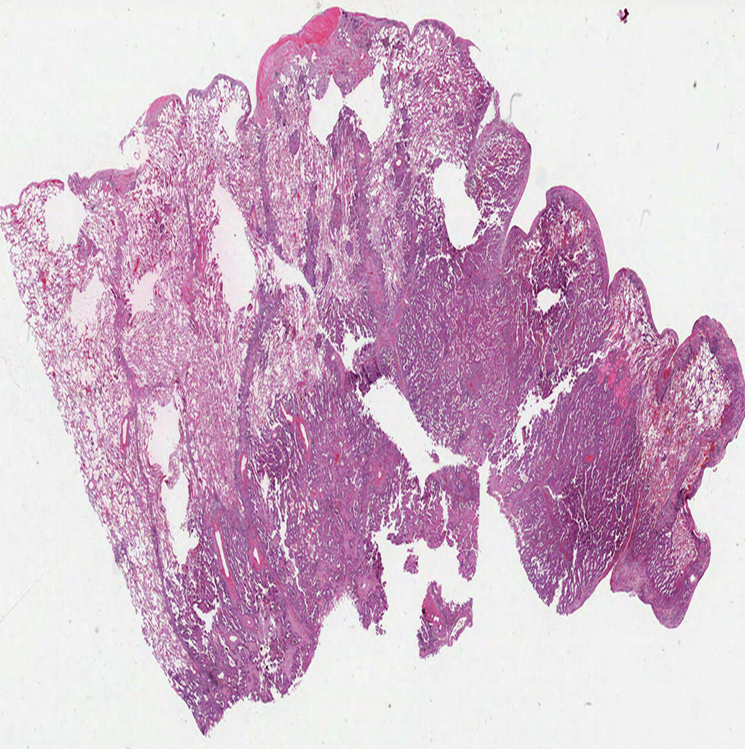} \\

\midrule
\textbf{Question} &
What is the TNM staging for a patient with a 7 cm tumor, several affected lymph nodes, but no distant metastasis?\\

\midrule
\textbf{WSI-LLaVA} &
The TNM staging for this patient is \textcolor{DarkGreen}{T3 N2 M0.}\\

\midrule
\textbf{Quilt-LLaVA} &
The TNM staging for this patient is \textcolor{DarkGreen}{T3 N2 M0.}\\

\midrule
\textbf{WSI-VQA} &
the tnm staging for this patient is \textcolor{red}{t2 n2 m0.}\\

\midrule
\textbf{GPT-4o} &
To determine the TNM staging, we need to consider the following components:- **T (Tumor size):** The tumor is 7 cm. This generally corresponds to T2 or T3, depending on the specific cancer type, but often T3 for many cancers. GPT-4o- **N(Nodes):** ¨ Several affected lymph nodes¨ typically indicates N1 or N2, depending on the exact number and extent of nodal involvement.- **M (Metastasis):** No distant metastasis is indicated as M0. Without specific details about the type of cancer, a general staging based on the information given could be \textcolor{red}{T3N1M0} or \textcolor{DarkGreen}{T3N2M0}. However, the exact stage can vary based on cancer type and guidelines, so consultation with specific cancer staging resources is advised for precise staging.\\
\midrule
\textbf{MLLM-HWSI} & 
\textcolor{DarkGreen}{According to TNM criteria, the 7 cm primary tumor meets the definition of T3 due to its size exceeding 5 cm.}
\textcolor{DarkGreen}{The description of “several affected lymph nodes” corresponds to N2, indicating regional nodal spread to multiple ipsilateral nodes.}
\textcolor{DarkGreen}{No evidence of distant metastasis defines M0.}
\textcolor{DarkGreen}{Therefore, the overall TNM staging for this patient is accurately classified as T3 N2 M0.}\\

\\

\midrule
\textbf{Ground Truth} &
TNM staging for this patient is T3 N2 M0.\\

\bottomrule
\end{tabularx}
}
\end{table*}

\section{Pre-training Details of MLLM-HWSI}
\label{training}
The pre-training of MLLM-HWSI is organized into three sequential stages: (\textbf{i}) hierarchical WSI–text alignment, (\textbf{ii}) hierarchical feature-space alignment, and (\textbf{iii}) task-specific instruction tuning.
Stages \textbf{I} and \textbf{II} utilize 9,642 WSI–caption pairs from the WSIBench dataset \cite{Liang2024_WSI-LLaVA} covering diverse cancer types, while Stage \textbf{III} employs 175,450 WSI-level VQA pairs from the same source for instruction fine-tuning.

Overall, the training process is divided into three stages, i.e., hierarchical WSI-text alignment, hierarchical feature space alignment, and task-specific instruction tuning.
In stage I and II, we used 9,642 WSIs-caption pairs from the WSIBench dataset \cite{Liang2024_WSI-LLaVA}.
In stage III, we used  175,450 WSI-level VQA pairs from the WSIBecnh dataset \cite{Liang2024_WSI-LLaVA}.
\\
\noindent \textbf{Stage I (Hierarchical WSI–Text Alignment).} In this stage, we align multi-scale WSI representations with their textual counterparts. 
The learning rate is set to $1 \times 10^{-3}$, and the batch size to 64.
Only the two-layer projection matrices responsible for vision–language alignment are optimized, while both the hierarchical encoders and the text encoder remain frozen. 
The model is trained for 50 epochs with a temperature coefficient of 0.02 to regulate the contrastive learning objective.
\\
\noindent \textbf{Stage II (Hierarchical Feature-Space Alignment).} During this phase, both the multi-scale visual encoder and the LLM remain frozen, and training focuses exclusively on refining the hierarchical projection layers to harmonize feature distributions across modalities. 
The learning rate is maintained at $1 \times 10^{-3}$, using a global batch size of 256 for one epoch.
The maximum input length is set to 2048 tokens, with no weight decay and a warmup ratio of 0.03 to ensure stable optimization.
\\
\noindent \textbf{Stage III (Instruction Fine-Tuning).} This stage enables multimodal reasoning by tuning the LLM jointly with the hierarchical projection layers while keeping the hierarchical encoder frozen. 
The learning rate is reduced to $2 \times 10^{-5}$, with a global batch size of 128 and a maximum sequence length of 2048. 
Weight decay remains 0, and the warmup ratio is fixed at 0.03. \
To achieve parameter-efficient adaptation, we apply LoRA (Low-Rank Adaptation) with a rank of 128 and $\alpha = 256$.
Training is performed using DeepSpeed ZeRO-3 for distributed optimization and BF16 precision with TensorFloat32 acceleration, improving computational efficiency while maintaining numerical stability.

\section{Computational Pathology Datasets}
\label{dataset}
To comprehensively evaluate MLLM-HWSI across a diverse range of CPath tasks, we employed multiple publicly available WSI datasets spanning classification, visual question answering (VQA), report generation, retrieval, and captioning benchmarks.

For \textbf{WSI classification}, including both zero-shot and linear probe evaluations, we used six standard benchmarks: BRACS \cite{brancati2022bracs}, PANDA \cite{Bulten2022_PANDA}, IMP-CRC \cite{neto2024interpretable}, TCGA-OT \cite{ding2024titan,mahmood2024tcgaot}, EBRAINS \cite{roetzer2022digital}, and UBC-Ocean \cite{UBC-OCEAN}. These datasets encompass a wide spectrum of organs, cancer subtypes, and histological grading systems, ensuring robust cross-domain generalization.

For the zero-shot \textbf{VQA} task, we adopted four multimodal benchmarks: WSI-Bench (4,119 pairs) \cite{Liang2024_WSI-LLaVA}, WSI-VQA (8,672 pairs) \cite{chen2025wsi}, SlideBench-VQA (BCNB) (7,247 pairs) \cite{chen2025slidechat}, and SlideBench-VQA (TCGA) (7,824 pairs) \cite{chen2025slidechat}. Together, these datasets evaluate the model’s ability to reason over morphological, diagnostic, and clinical questions at the slide level.

For \textbf{report generation}, we used the WSI-Bench (208 WSI–report pairs) \cite{liang2025wsillavamultimodallargelanguage} and HistGen (700 pairs) \cite{guo2024histgen} datasets, both curated to assess automatic report synthesis grounded in morphological evidence.

For the \textbf{WSI retrieval} task, we evaluated on TCGA-OT \cite{ding2024titan,mahmood2024tcgaot}, EBRAINS \cite{roetzer2022digital}, and IMP-CRC \cite{neto2024interpretable}, enabling assessment of large-scale visual similarity retrieval in diagnostic contexts.

For \textbf{cross-modal retrieval}, we utilized the TCGA Reports dataset \cite{weinstein2013cancer,ding2024titan}, which links WSIs with associated clinical and textual records to evaluate bidirectional alignment between visual and textual representations.

Finally, for \textbf{caption generation}, we used the SlideBench dataset \cite{chen2025slidechat}, designed for producing concise, pathology-grounded descriptions of WSIs.

Collectively, these datasets provide a comprehensive evaluation suite for assessing MLLM-HWSI’s performance across diagnostic interpretation, reasoning, and language grounding tasks in computational pathology.

\noindent \textbf{1. BRACS (7 classes)} \cite{brancati2022bracs} consists of 547 H\&E FFPE WSIs of breast tumors (benign, atypical, and malignant) collected from 189 patients. The cases are annotated at two levels: a coarse-grained level of three classes (benign tumors: 265, atypical tumors: 89, malignant tumors: 193) and a fine-grained level of seven subtypes (including invasive carcinoma, ductal carcinoma in situ, and various benign/atypical hyperplasias). The dataset is divided into five label-stratified, patient-level splits using a 60:20:20 ratio (approx. 302:94:151 slides) for training, validation, and testing.
\\

\noindent \textbf{2. UBC-Ocean (5 Classes)} \cite{UBC-OCEAN} comprises 538 WSIs, with 527 meeting foreground tissue criteria, for ovarian cancer subtyping. The dataset covers five distinct subtypes: Clear Cell (CC), Endometrioid (EC), High-Grade Serous Carcinoma (HGSC), Low-Grade Serous Carcinoma (LGSC), and Mucinous Carcinoma (MC). The dataset is divided in a stratified fashion into train:validation:test sets with approximately 369:52:106 WSIs, respectively.
\\

\noindent \textbf{3. TCGA-OT (46 Classes)} \cite{ding2024titan,mahmood2024tcgaot} is a pan-cancer subtyping dataset derived from TCGA, consisting of 11,186 H\&E FFPE diagnostic histopathology WSIs of primary tumors. All WSIs are classified into 46 distinct cancer types based on the OncoTree classification system, with each class represented by at least 50 samples. Slides were rigorously curated by excluding frozen tissues, metastatic/recurrent tumors, and slides lacking magnification or tumor tissue. The dataset is split into training, validation, and test folds of 8,226:1,612:1,348 samples, respectively, while ensuring all slides from the same source site remain within a single split.
\\

\noindent \textbf{4. EBRAINS (30 classes)} dataset \cite{roetzer2022digital} features H\&E-stained whole-slide images (WSIs) of brain tissue sourced from The Digital Brain Tumour Atlas. For our study, we utilized a subset of 2,319 WSIs (out of 3,114 total), mirroring the selection process used for the CONCH dataset \cite{lu2023visual}. This defined a 30-class fine-grained brain tumor subtyping task, including only diagnostic labels with at least 30 slides. We established the WSI counts per class to match those in CONCH. For the supervised task, the 2,319 slides were split 50\%-25\%-25\% into training (1,151 slides), validation (595 slides), and testing (573 slides). This 573-slide testing split was also used as the zero-shot test set.
\\

\noindent \textbf{5. PANDA (6 classes)} is the International Society of Urological Pathology (ISUP) grading task derived from the PANDA challenge \cite{Bulten2022_PANDA}. This dataset comprises prostate cancer core needle biopsies. We utilized a subset of 9,555 Whole Slide Images (WSIs) after excluding noisy labels from the original 10,616 slides. These 9,555 slides are distributed across the six ISUP grades as follows: Grade 0 (2,603), Grade 1 (2,399), Grade 2 (1,209), Grade 3 (1,118), Grade 4 (1,124), and Grade 5 (1,102). For experiments, the dataset was partitioned into standard 80\% training, 10\% validation, and 10\% test sets (7,647:954:954 WSIs).
\\

\noindent \textbf{6. IMP-CRC (3 Classes)} \cite{neto2024interpretable} is a colorectal cancer dataset containing 5,333 H\&E FFPE biopsy and polypectomy WSIs from the IMP Diagnostics laboratory. Cases are classified into three distinct categories: Non-neoplastic (847 slides), Low-grade lesions (2847 slides) which include conventional adenomas with low-grade dysplasia, and High-grade lesions (1639 slides) encompassing conventional adenomas with high-grade dysplasia, intramucosal carcinomas, and invasive adenocarcinomas. The dataset is label-stratified and split into train:validation:test sets using a 60:20:20 ratio, resulting in 3546:887:900 slides, respectively.
\\

\noindent \textbf{7. WSI-Bench} \cite{Liang2024_WSI-LLaVA} is a large-scale VQA dataset specifically designed for WSIs. It contains a total of 179,569 VQA pairs. The training set comprises 175,450 pairs across 9,642 WSIs (122,133 open-ended and 53,317 closed-ended questions). The test set consists of 4,119 VQA pairs from 208 WSIs (2,838 open-ended and 1,281 closed-ended questions). Additionally, a specific subset of 208 VQA pairs is dedicated to report generation.
\\

\noindent \textbf{8. WSI-VQA} dataset \cite{Chen2025_WSI-VQA} contains 977 whole-slide images (WSIs), which are paired with a total of 8,672 question-and-answer (QA) pairs. On average, this amounts to approximately 8.9 QA pairs per WSI. The QA pairs are composed of 4,535 close-ended questions and 4,137 open-ended questions.
\\

\noindent \textbf{9. SlideBench-VQA (BNCB)} \cite{chen2025slidechat} is a dataset comprising 7,247 Visual Question Answering (VQA) pairs derived from 1,058 patients. Its primary purpose is to evaluate the zero-shot generalization capability of models like SlideChat across seven distinct classification tasks.
\\

\noindent \textbf{10. SlideBench-VQA (TCGA)} \cite{chen2025slidechat} is a VQA dataset specifically focused on WSIs sourced from The Cancer Genome Atlas (TCGA). The dataset comprises 7,827 VQA pairs, which cover 13 distinct WSI categories.
The 2451 overlapping samples of SlideBench-VQA (test split) with WSI-Bench were not used during training. 
All evaluations were performed on held-out test splits.
Our zero-shot results, therefore, reflect generalization to unseen WSIs.
\\

\noindent \textbf{11. HistGen-Report} \cite{guo2024histgen} is a WSI dataset designed for report generation. It comprises 7,753 WSI-report pairs sourced from the TCGA platform. The diagnostic reports were subsequently refined using large language models to ensure high quality, coherence, and diagnostic relevance.
\\


\noindent \textbf{12. TCGA-Reports} \cite{weinstein2013cancer,ding2024titan} is a dataset containing pathology reports sourced from The Cancer Genome Atlas (TCGA) data portal. The dataset was compiled from 11,108 pathology report PDFs, corresponding to 11,010 patients.
\\

\end{document}